\documentclass{article}

\usepackage{microtype}
\usepackage{graphicx}
\usepackage{subfigure}
\usepackage{booktabs} 

\usepackage{hyperref}

\usepackage{graphicx} 
\usepackage{subfigure}
\usepackage{amsmath}
\usepackage{bm}
\usepackage{multirow}
\usepackage{amssymb}
\usepackage{subfigure}
\usepackage{wrapfig}
\usepackage{enumerate}

\usepackage[accepted]{icml2020}

\icmltitlerunning{Deep Multi-Task Learning via Generalized Tensor Trace Norm}

\begin{document}

\twocolumn[
\icmltitle{Deep Multi-Task Learning via Generalized Tensor Trace Norm}



\icmlsetsymbol{equal}{*}

\begin{icmlauthorlist}
\icmlauthor{Yi Zhang}{sustech}
\icmlauthor{Yu Zhang}{sustech}
\icmlauthor{Wei Wang}{nju}
\end{icmlauthorlist}

\icmlaffiliation{sustech}{Department of Computer Science and Engineering, Southern University of Science and Technology, China}
\icmlaffiliation{nju}{Department of Computer Science and Technology, Nanjing University, China}

\icmlcorrespondingauthor{Yu Zhang}{yu.zhang.ust@gmail.com}

\icmlkeywords{Machine Learning, ICML}

\vskip 0.3in
]



\printAffiliationsAndNotice

\begin{abstract}

The trace norm is widely used in multi-task learning as it can discover low-rank structures among tasks in terms of model parameters. Nowadays, with the emerging of big datasets and the popularity of deep learning techniques, tensor trace norms have been used for deep multi-task models. However, existing tensor trace norms cannot discover all the low-rank structures and they require users to manually determine the importance of their components. To solve those two issues together, in this paper, we propose a Generalized Tensor Trace Norm (GTTN). The GTTN is defined as a convex combination of matrix trace norms of all possible tensor flattenings and hence it can discover all the possible low-rank structures. In the induced objective function, we will learn combination coefficients in the GTTN to automatically determine the importance. Experiments on real-world datasets demonstrate the effectiveness of the proposed GTTN.

\end{abstract}

\newtheorem{theorem}{Theorem}
\newtheorem{lemma}{Lemma}
\newtheorem{definition}{Definition}
\newtheorem{remark}{Remark}
\newtheorem{corollary}{Corollary}

\newcommand{\vertiii}[1]{{\left\vert\kern-0.25ex\left\vert\kern-0.25ex\left\vert #1
    \right\vert\kern-0.25ex\right\vert\kern-0.25ex\right\vert}}

\section{Introduction}

Given multiple related learning tasks, multi-task learning \cite{caruana97,zy17b} aims to exploit useful information contained in them to help improve the performance of all the tasks. Multi-task learning has been applied to many application areas, including computer vision, natural language processing, speech recognition and so on. Over past decades, many multi-task learning models have been devised to learn such useful information shared by all the tasks. As reviewed in \cite{zy17b}, multi-task learning models can be categorized into six classes, including the feature transformation approach \cite{aep06,msgh16}, feature selection approach \cite{otj06,lpz09,ls12}, low-rank approach \cite{ptjy10,hz16,yh17}, task clustering approach \cite{xlck07,jbv08,kd12,hz15a}, task relation learning approach \cite{bcw07,zy10a,lcwy17,zwy18}, and decomposition approach \cite{jrsr10,cly10,zw13,hz15b}.

Among those approaches, the low-rank approach is effective to identify low-rank model parameters. When model parameters of a task can be organized in a vector corresponding to for example binary classification tasks or regression tasks on vectorized data, the matrix trace norm or its variants is used as a regularizer on the parameter matrix, each of whose columns stores parameters for a task, to identify the low-rank structure among tasks. Nowadays with the collection of complex data and the popularity of deep learning techniques, each data point can be represented as a tensor (e.g., images) and each learning task becomes complex, e.g., multi-class classification tasks. In this case, the parameters of all the tasks are stored in a tensor, making the matrix trace norm not applicable, and instead tensor trace norms \cite{rabp13,wst14,yh17} are used to learn low-rank parameters in the parameter tensor for multi-task learning.

Different from the matrix trace norm which has a unique definition, the tensor trace norm has many variants as the tensor rank has multiple definitions. Here we focus on overlapped tensor trace norms which equals the sum of the matrix trace norm of several tensor flattenings of the tensor. An overlapped tensor trace norm relies on the way to do the tensor flattening. For example, the Tucker trace norm \cite{tucker66} conducts the tensor flattening along each axis in the tensor and the Tensor-Train (TT) trace norm \cite{oseledets11} does it along successive axes starting from the first one. There are two limitation in the existing tensor trace norms. Firstly, for a $p$-way tensor, we can see that there are $2^p-2$ possible tensor flattenings but existing overlapped tensor trace norms only utilize a subset of them, making them fail to capture all the low-rank structures in the parameter tensor. Another limitation of existing tensor trace norms is that all the tensor flattenings used in a tensor trace norm are assumed to be equally important, which is suboptimal to the performance.

In this paper, to overcome the two aforementioned limitations of existing overlapped tensor trace norms, we propose a Generalized Tensor Trace Norm (GTTN). The GTTN exploits all possible tensor flattenings and it is defined as the convex sum of matrix trace norms of all possible tensor flattenings. In this way, the GTTN can capture all the low-rank structures in the parameter tensor and hance overcome the first limitation. Moreover, to alleviate the second limitation, we treat combination coefficients in the GTTN as variables and propose an objective function to learn them from data. Another advantage of learning combination coefficients is that it can show the importance of some axes, which can improve the interpretability of the learning model and give us insights for the problem under investigation. To obtain a full understanding of the GTTN, we study properties of the proposed GTTN. For example, the number of tensor flattenings with distinct matrix trace norms is proved to be $2^{p-1}-1$ and so when $p\le 5$ we encountered in most problems, such number is not so large that the computational complexity is comparable to existing tensor trace norms. We also analyze the dual norm of the GTTN and give a generalization bound. Extensive experiments on real-world datasets demonstrate the effectiveness of the proposed GTTN.

\section{Existing Tensor Trace Norms}


In multi-task learning, trace norms are widely used as the regularization to learn a low-rank structure among model parameters of all the tasks as minimizing the trace norm will enforce some singular values to approach zero. When both a data point and model parameters of a task are represented in vectorized forms in regression tasks or binary classification tasks, the matrix trace norm can be used and it is defined as $\|\mathbf{W}\|_*=\sum_{i}\sigma_i(\mathbf{W})$ with each column of the parameter matrix $\mathbf{W}$ storing the parameter vector of the corresponding task and $\sigma_i(\mathbf{W})$ denoting the $i$th largest singular value of $\mathbf{W}$. Regularizing $\mathbf{W}$ with $\|\mathbf{W}\|_*$ will make $\mathbf{W}$ tend to be low-rank, which leads to the linear dependency among parameter vectors of different tasks and reflects the relatedness among tasks in terms of model parameters.

Nowadays, the data such as images can be represented in a matrix or tensor form in the raw representation (e.g., pixel-based representation) and transformed representation after for example convolutional operations. Moreover, each task becomes more complex, for example, a multi-class classification task. In those cases, parameters of all the tasks can be organized as a $p$-way tensor ($p\ge 3$), e.g., $\mathcal{W}\in\mathbb{R}^{d_1\times\ldots\times d_p}$. That is, for multi-class classification tasks, when $p$ equals 3, $d_1$ denotes the number of hidden units in the last hidden layer, $d_2$ can represent the number of classes, and $d_3$ can be the number of tasks. In such cases, the matrix trace norm is no longer applicable and instead tensor trace norms are investigated.

According to \cite{ts13}, tensor trace norms can be classified into two categories, including overlapped tensor trace norms and latent tensor trace norms. An overlapped tensor trace norm transforms a tensor into matrices in different ways and compute the sum of the matrix trace norm of different transformed matrices. A latent tensor trace norm decomposes the tensor into multiple latent tensors and then compute the sum of the matrix trace norm of matrices which are transformed from the latent tensors. Deep multi-task learning mainly uses the overlapped tensor trace norm, which is the focus of our study.

As reviewed in \cite{yh17}, three tensor trace norms belonging to the overlapped tensor trace norm are used in deep multi-task learning, including the Tucker trace norm, TT trace norm, and Last Axis Flattening (LAF) trace norm. In the following, we will review those three tensor trace norms.

\subsection{Tucker Trace Norm}

Based on the Tucker decomposition \cite{tucker66}, the Tucker trace norm for a tensor $\mathcal{W}\in\mathbb{R}^{d_1\times\ldots\times d_p}$ can be defined as
\begin{equation}
\vertiii{\mathcal{W}}_*=\sum_{i=1}^p\alpha_i\|\mathcal{W}_{(i)}\|_*,\label{def_Tucker_trace_norm}
\end{equation}
where $[p]$ denotes a set of positive integers no larger than $p$, $\mathrm{permute}(\mathcal{W},\mathbf{s})$ permutes the tensor $\mathcal{W}$ along axis indices in $\mathbf{s}$ that is a permutation of $[p]$, $\mathrm{reshape}(\mathcal{W},\mathbf{a})$ reshapes the tensor $\mathcal{W}$ with the new size stored in a vector $\mathbf{a}$, $\mathcal{W}_{(i)}:=\mathrm{reshape}(\mathrm{permute}(\mathcal{W},[i,1,\ldots,i-1,i+1,\ldots,p]),[d_i,\prod_{j\ne i}d_j])$ is the mode-$i$ tensor flattening to transform $\mathcal{W}$ to a matrix along the $i$th axis, and $\alpha_i$ denotes the weight for the model-$i$ flattening. To control the scale of $\{\alpha_i\}$, here $\{\alpha_i\}$ are required to satisfy that $\alpha_i\ge 0$ and $\sum_{i=1}^p\alpha_i=1$. Based on Eq. (\ref{def_Tucker_trace_norm}), we can see that the Tucker trace norm is a convex combination of matrix trace norms of tensor flattening along each axis, where $\alpha_i$ controls the importance of the mode-$i$ tensor flattening. Without a priori information, different tensor flattenings are usually assumed to have equal importance by setting $\alpha_i$ to be $\frac{1}{p}$.

Besides being used in deep multi-task learning, the Tucker trace norm has been adopted in multilinear multi-task learning \cite{rabp13,wst14} which assumes the existence of multi-modal structures contained in multi-task learning problems.

\subsection{TT Trace Norm}

Based on the tensor-train decomposition \cite{oseledets11}, the TT trace norm for a tensor $\mathcal{W}\in\mathbb{R}^{d_1\times\ldots\times d_p}$ can be defined as
\begin{equation}
\vertiii{\mathcal{W}}_*=\sum_{i=1}^{p-1}\alpha_i\|\mathcal{W}_{[i]}\|_*,\label{def_TT_trace_norm}
\end{equation}
where $\mathcal{W}_{[i]}=\mathrm{reshape}(\mathcal{W},[\prod_{j=1}^id_j,\prod_{j=i+1}^pd_j])$ and $\alpha_i$ denotes a nonnegative weight. Different from the mode-$i$ tensor flattening $\mathcal{W}_{(i)}$, $\mathcal{W}_{[i]}$ unfolds the tensor to a matrix along the first $i$ axes. Similar to the Tucker trace norm, $\{\alpha_i\}$ are assumed to satisfy that $\alpha_i\ge 0$ and $\sum_{i=1}^{p-1}\alpha_i=1$. Usually $\alpha_i$ is set by users to be $\frac{1}{p-1}$ if there is no additional information for the importance of each term in Eq. (\ref{def_TT_trace_norm}).

\subsection{LAF Trace Norm}

The LAF trace norm for a tensor $\mathcal{W}\in\mathbb{R}^{d_1\times\ldots\times d_p}$ can be defined as
\begin{equation}
\vertiii{\mathcal{W}}_*=\|\mathcal{W}_{(p)}\|_*.\label{def_LAF_trace_norm}
\end{equation}
The last axis in $\mathcal{W}$ is the task axis and hence the LAF trace norm is equivalent to place the matrix trace norm on $\mathcal{W}_{(p)}$ each of whose rows stores model parameters of each task. Compared with the Tucker trace norm in Eq. (\ref{def_Tucker_trace_norm}), the LAF trace norm can be viewed as a special case of the Tucker trace norm where $\alpha_p$ equals 1 and other $\alpha_i$'s ($i\ne p$) are equal to 0.

Given a tensor trace norm, the objective function of a deep multi-task model can be formulated as\footnote{Here for simplicity, we assume the tensor trace norm regularization is placed on only one $\mathcal{W}$. This formulation can easily be extended to multiple $\mathcal{W}$'s with the tensor trace norm regularization.}
\begin{equation}
\min_{\bm{\Theta}}\sum_{i=1}^m\frac{1}{n_i}\sum_{j=1}^{n_i}l(f_i(\mathbf{x}^i_j;\bm{\Theta}),y^i_j)+\lambda\vertiii{\mathcal{W}}_*,
\label{obj_DMTL}
\end{equation}
where $m$ denotes the number of tasks, $n_i$ denotes the number of training data points in the $i$th task, $\mathbf{x}^i_j$ denotes the $j$th data point in the $i$th task, $y^i_j$ denotes the label of $\mathbf{x}^i_j$, $f_i(\cdot;\bm{\Theta})$ denotes a learning function for the $i$th task given a deep multi-task neural network parameterized by $\bm{\Theta}$, $l(\cdot,\cdot)$ denotes a loss function such as the cross-entropy loss for classification tasks and the square loss for regression tasks, $\mathcal{W}$ denotes a part of $\bm{\Theta}$ that is regularized by a tensor trace norm, and $\lambda$ is a regularization parameter. In problem (\ref{obj_DMTL}), the tensor trace norm can be the Tucker trace norm, or the TT trace norm, or the LAF trace norm.

\section{Generalized Tensor Trace Norm}

In this section, we first analyze existing tensor trace norms and then present the proposed generalized tensor trace norm as well as the optimization and generalization bound.

\subsection{Analysis on Existing Tensor Trace Norms}

As introduced in the previous section, we can see that overlapped tensor trace norms rely on different ways of tensor flattening. For example, the Tucker trace norm reshapes the tensor along each axis and the LAF trace norm focuses on the last axis, while the TT trace norm reshapes the tensor by combining the first several axes. Given the physical meaning of each axis, the LAF trace norm only considers the inter-task low-rank structure among tasks, but differently both the Tucker and TT trace norms consider  not  only the inter-task low-rank structure among tasks but  also the intra-task low-rank structure among, for example, features. In this sense, the Tucker and TT trace norms seems to be superior to the LAF trace norm.

For overlapped tensor trace norms like the Tucker and TT trace norms, there are two important issues.
\begin{enumerate}[1)]
\item How to choose the ways of tensor flattening?
\item How to determine the importance of of different ways of tensor flattening?
\end{enumerate}

For the first issue, the Tucker trace norm chooses to reshape along each axis while the TT trace norm combines the first several axes together to do the tensor flattening. Different ways of tensor flattening encode the belief on the existence of the low-rank structure in $\mathcal{W}$. So the Tucker trace norm assumes that the low-rank structure exists in each axis while the TT trace norm considers the combinations of the first several axes have low-rank structure. However, those models may fail when such assumptions do not hold.

For the second issue, current models usually assume the equal importance of different ways of tensor flattening, which is reflected in the equal value of $\{\alpha_i\}$. Intuitively, different ways of tensor flattening should have different degrees in terms of the low-rank structure and hence $\{\alpha_i\}$ should be different from each other. In this sense, $\{\alpha_i\}$ with an equal value incur the suboptimal performance.


\subsection{GTTN}

To solve the above two issues together, we propose the generalized tensor trace norm.

For the first issue, since for most problems we do not know which ways of tensor flattening are helpful to learn the low-rank structure, we can try all possible ways of tensor flattening. To mathematically define this, we define $\mathcal{W}_{\{\mathbf{s}\}}$ as
\begin{equation*}
\mathcal{W}_{\{\mathbf{s}\}}=\mathrm{reshape}\Big(\mathrm{permute}(\mathcal{W},[\mathbf{s},\neg\mathbf{s}]),
\Big[\prod_{i\in\mathbf{s}}d_i,\prod_{j\in\neg\mathbf{s}}d_j\Big]\Big),
\end{equation*}
where $\mathbf{s}$ is a nonempty subset of $[p]$ (i.e., $\mathbf{s}\subset[p]$) and $\neg\mathbf{s}$ denotes the complement of $\mathbf{s}$ with respect to $[p]$ (i.e., $\neg\mathbf{s}=[p]-\mathbf{s}$). So $\mathcal{W}_{\{\mathbf{s}\}}$ is a tensor flattening to a matrix with a dimension corresponding to axis indices in $\mathbf{s}$ and the other to axis indices in $\neg\mathbf{s}$. When $\mathbf{s}=\{i\}$ contains only one element, $\mathcal{W}_{\{\mathbf{s}\}}$ becomes $\mathcal{W}_{(i)}$, the mode-$i$ tensor flattening used in the Tucker trace norm. When $\mathbf{s}=[i]$, $\mathcal{W}_{\{\mathbf{s}\}}$ becomes $\mathcal{W}_{[i]}$ that is used in the TT trace norm. Moreover, this new tensor flattening can be viewed as a generalization of $\mathcal{W}_{(i)}$ and $\mathcal{W}_{[i]}$ as $\mathbf{s}$ can contain more than one element, which is more general than $\mathcal{W}_{(i)}$, and it does not require that elements in $\mathbf{s}$ should be successive integers from 1, which is more general than $\mathcal{W}_{[i]}$.

As we aim to consider all possible ways of tensor flattening, similar to the Tucker and TT trace norms, we define the GTTN as
\begin{equation}
\vertiii{\mathcal{W}}_*=\sum_{\mathbf{s}\subset[p],\mathbf{s}\ne\emptyset}\alpha_{\mathbf{s}}\|\mathcal{W}_{\{\mathbf{s}\}}\|_*,\label{GTTN}
\end{equation}
where $\mathbf{s}$ is also used as a subscript to index the corresponding weight for $\|\mathcal{W}_{\{\mathbf{s}\}}\|_*$, $\bm{\alpha}$ denotes the set of $\alpha_{\mathbf{s}}$'s, $\mathcal{C}_{\bm{\alpha}}=\{\bm{\alpha}|\alpha_{\mathbf{s}}\ge 0$ and $\sum_{\mathbf{s}\subset[p]}\alpha_{\mathbf{s}}=1\}$ defines a constraint set for $\bm{\alpha}$. Then based on the GTTN, we can solve the first issue to some extent as it can discover all the low-rank structures by considering all possible ways of tensor flattening with appropriate settings of $\bm{\alpha}$.

\begin{figure}[t]
\vskip -0.05in
\begin{center}
\includegraphics[width=1\linewidth]{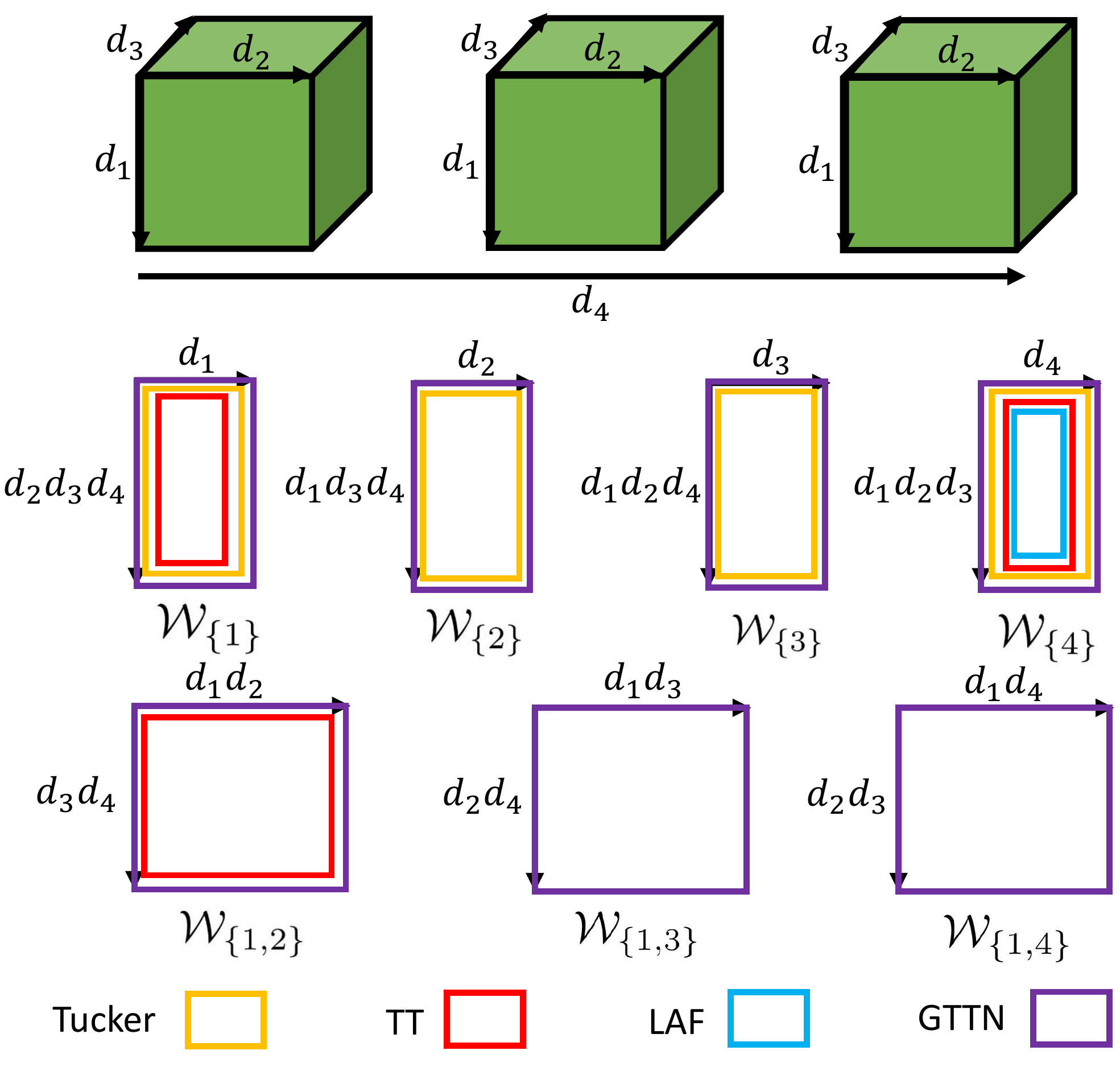}
\end{center}
\vskip -0.2in
\caption{Comparison among the Tucker trace norm, TT trace norm, LAF trace norm, and GTTN. At the top, there is a four-way tensor where each cube is a slice along the last axis. The 7 matrices denotes all the possible tensor flattenings. If a tenor flattening is used by the Tucker trace norm, it will have an orange rectangle.  If a tenor flattening is used by the TT trace norm, it will have a red rectangle. If a tenor flattening is used by the LAF trace norm, it will have a blue rectangle. If a tenor flattening is used by the GTTN, it will have a purple rectangle.}
\label{fig_GTTN}
\vskip -0.15in
\end{figure}

In Figure \ref{fig_GTTN}, we show the difference among the Tucker trace norm, TT trace norm, LAF trace norm and GTTN for a four-way tensor at the top. In the bottom of Figure \ref{fig_GTTN}, we can see that there are seven possible tensor flattenings. The Tucker trace norm uses $\mathcal{W}_{\{1\}}$, $\mathcal{W}_{\{2\}}$, $\mathcal{W}_{\{3\}}$, and $\mathcal{W}_{\{4\}}$. The TT trace norm relies on $\mathcal{W}_{\{1\}}$, $\mathcal{W}_{\{1,2\}}$, and $\mathcal{W}_{\{1,2,3\}}$. The LAF trace norm only contains $\mathcal{W}_{\{4\}}$. The calculation of the GTTN is based on all the seven tensor flattenings. From this example, we can see that the union of tensor flattenings used in the Tucker, TT, and LAF trace norms cannot cover all the possible ones and the GTTN utilizes some additional tensor flattening (e.g., $\mathcal{W}_{\{1,3\}}$ and $\mathcal{W}_{\{1,4\}}$). In this sense, the GTTN can discover more low-rank structures than existing tensor trace norms.

For the number of distinct summands in the right-hand side of Eq. (\ref{GTTN}), we have the following theorem.\footnote{All the proofs are put in the appendix.}

\begin{theorem}\label{theorem_num_summands}
The right-hand side of Eq. (\ref{GTTN}) has $2^{p-1}-1$ distinct summands.
\end{theorem}

As shown in the proof of Theorem \ref{theorem_num_summands}, $\mathcal{W}_{\{\mathbf{s}\}}$ and $\mathcal{W}_{\{\neg\mathbf{s}\}}$ are transpose matrices to each other with equal matrix trace norm and we can eliminate one of them to reduce the computational cost. For notational simplicity, we do not explicitly do the elimination in the formulation but in computation, we did do that. In problems we encounter, $p$ is at most $5$ and so the GTTN has at most 15 distinct summands. So the number of distinct summands are not so large, making the optimization efficient.

Similar to the Tucker and TT trace norms, GTTN defined in Eq. (\ref{GTTN}) still faces the second issue. Here to solve the second issue, we view $\bm{\alpha}$ as variables to be optimized and based on Eq. (\ref{GTTN}), the objective function of a deep multi-task model based on GTTN is formulated as
\begin{equation}
\min_{\bm{\Theta},\bm{\alpha}\in\mathcal{C}_{\bm{\alpha}}}\sum_{i=1}^m\frac{1}{n_i}\sum_{j=1}^{n_i}l(f_i(\mathbf{x}^i_j;\bm{\Theta}),y^i_j)
+\lambda\vertiii{\mathcal{W}}_*.
\label{obj_GTTN}
\end{equation}
Compared with problem (\ref{obj_DMTL}), we can see two differences. Firstly, the regularization terms in two problems are different. Secondly, problem (\ref{obj_GTTN}) treat $\bm{\alpha}$ as variables to be optimized but the corresponding entities are constants which are set by users.

In the following theorem, we can simplify problem (\ref{obj_GTTN}) by eliminating $\bm{\alpha}$.

\begin{theorem}\label{theorem_obj_GTTN_reformulation}
Problem (\ref{obj_GTTN}) is equivalent to
\begin{equation}
\min_{\bm{\Theta}}\sum_{i=1}^m\frac{1}{n_i}\sum_{j=1}^{n_i}l(f_i(\mathbf{x}^i_j;\bm{\Theta}),y^i_j)
+\lambda\min_{{\mathbf{s}\subset[p]\atop\mathbf{s}\ne\emptyset}}\|\mathcal{W}_{\{\mathbf{s}\}}\|_*,\label{obj_GTTN_2}
\end{equation}
\end{theorem}

According to problem (\ref{obj_GTTN_2}), learning $\bm{\alpha}$ will tend to choosing a tensor flattening with the minimal matrix trace norm.

\subsection{Optimization}

Even though problem (\ref{obj_GTTN_2}) is equivalent to problem (\ref{obj_GTTN}), in numerical optimization, we choose problem (\ref{obj_GTTN}) as the objective function to be optimized. One reason is that problem (\ref{obj_GTTN_2}), which involves the minimum of matrix trace norms, is more complicated than problem (\ref{obj_GTTN}) to be optimized. Another reason is that the learned $\bm{\alpha}$ in problem (\ref{obj_GTTN}) can visualize the importance of each tensor flattening, which can improve the interpretability of the learning model.

Since problem (\ref{obj_GTTN}) is designed for deep neural networks, the Stochastic Gradient Descent (SGD) technique is the first choice for optimization. However, problem (\ref{obj_GTTN}) is a constrained optimization problem, making SGD techniques not directly applicable. The constraints in problem (\ref{obj_GTTN}) constrain $\bm{\alpha}$ to form a $(p-1)$-dimensional simplex. To convert problem (\ref{obj_GTTN}) to an unconstrained problem that can be optimized by SGD, we reparameterize each $\alpha_{\mathbf{s}}$ as
\begin{equation*}
\alpha_{\mathbf{s}}=\frac{\exp\{\beta_{\mathbf{s}}\}}{\sum_{\mathbf{t}\subset[p],\mathbf{t}\ne\emptyset}\exp\{\beta_{\mathbf{t}}\}}.
\end{equation*}
With such reparameterization, problem (\ref{obj_GTTN}) can be reformulated as
{\small
\begin{equation}
\min_{\bm{\Theta},\bm{\beta}}\sum_{i=1}^m\frac{1}{n_i}\sum_{j=1}^{n_i}l(f_i(\mathbf{x}^i_j;\bm{\Theta}),y^i_j)
+\frac{\lambda\sum_{\mathbf{s}\subset[p]\atop\mathbf{s}\ne\emptyset}\exp\{\beta_{\mathbf{s}}\}\|\mathcal{W}_{\{\mathbf{s}\}}\|_*}
{\sum_{\mathbf{t}\subset[p],\mathbf{t}\ne\emptyset}\exp\{\beta_{\mathbf{t}}\}}.\label{obj_GTTN_3}
\end{equation}
}\noindent
For each parameter $\theta\in\bm{\Theta}-\mathcal{W}$, its gradient can be computed based on the first term in the objective function of problem (\ref{obj_GTTN_3}). For each $\beta_{\mathbf{s}}$, its gradient can be computed as
\begin{eqnarray*}
\frac{\partial h}{\partial \beta_{\mathbf{s}}}&=&
-\frac{\lambda\exp\{\beta_{\mathbf{s}}\}\sum_{\mathbf{t}\subset[p]\atop\mathbf{t}\ne\emptyset}
\exp\{\beta_{\mathbf{t}}\}\|\mathcal{W}_{\{\mathbf{t}\}}\|_*}
{\left(\sum_{\mathbf{t}\subset[p],\mathbf{t}\ne\emptyset}\exp\{\beta_{\mathbf{t}}\}\right)^2}\\
&&+\frac{\lambda\exp\{\beta_{\mathbf{s}}\}\|\mathcal{W}_{\{\mathbf{s}\}}\|_*}
{\sum_{\mathbf{t}\subset[p],\mathbf{t}\ne\emptyset}\exp\{\beta_{\mathbf{t}}\}}.
\end{eqnarray*}
For $\mathcal{W}$, the computation of its gradient comes from both terms in the objective function of problem (\ref{obj_GTTN_3}). The first term is the conventional training loss and the second term involves the matrix trace norm which is non-differentiable. According to \cite{watson92}, we can compute the subgradient instead, that is, $\frac{\partial \|\mathbf{X}\|_*}{\partial \mathbf{X}}=\mathbf{U}\mathbf{V}^T$ where $\mathbf{X}=\mathbf{U}\bm{\Sigma}\mathbf{V}^T$ denotes the singular value decomposition of a matrix $\mathbf{X}$.

\subsection{Generalization Bound}

For the GTTN defined in Eq. (\ref{GTTN}), we can derive its dual norm in the following theorem.

\begin{theorem}\label{theorem_GTTN_dual_norm}
The dual norm of the GTTN defined in Eq. (\ref{GTTN}) is defined as
\begin{equation*}
\vertiii{\mathcal{W}}_{*^{\star}}=
\min_{\sum_{\mathbf{s}\ne\emptyset\atop\mathbf{s}\subset[p]}\alpha_{\mathbf{s}}\mathcal{Y}^{(\mathbf{s})}=\mathcal{W}}
\max_{\mathbf{s}\ne\emptyset\atop\mathbf{s}\subset[p]}\|\mathcal{Y}^{(\mathbf{s})}_{\{\mathbf{s}\}}\|_{\infty},
\end{equation*}
where $\mathcal{Y}^{(\mathbf{s})}$ is a variable indexed by $\mathbf{s}$ and $\|\cdot\|_{\infty}$ denotes the spectral norm of a matrix that is equal to the maximum singular value.
\end{theorem}

Without loss of generality, here we assume $\bm{\Theta}=\mathcal{W}$ which can simplify the analysis. We rewrite problem (\ref{obj_GTTN}) into an equivalent formulation as
\begin{equation}
\min_{\mathcal{W}}\sum_{i=1}^m\frac{1}{n_i}\sum_{j=1}^{n_i}l(f_i(\mathbf{x}^i_j;\mathcal{W}),y^i_j)\ \mathrm{s.t.}\ \vertiii{\mathcal{W}}_*\le\gamma,\label{obj_GTTN_4}
\end{equation}
where $\bm{\alpha}$ is assumed to be fixed to show its impact to the bound. Here each data point is a tensor and binary classification tasks are considered,\footnote{The analysis is easy to extend to regression tasks and multi-class classification tasks.} implying that $\mathcal{W}\in\mathbb{R}^{d_1\times\ldots\times d_{p-1}\times m}$ and $\mathbf{x}^i_j\in\mathbb{R}^{d_1\times\ldots\times d_{p-1}}$. The learning function for each task is a linear function defined as $f_i(\mathbf{x};\mathcal{W})=\langle\mathcal{W}_i,\mathbf{x}\rangle$, where $\langle\cdot,\cdot\rangle$ denotes the inner product between two tensors with equal size and $\mathcal{W}_i$ denotes the $i$th slice along the last axis which is the task axis. For simplicity, different tasks are assumed to have the same number of data points, i.e., $n_i$ equals $n_0$ for $i=1,\ldots,m$. It is very easy to extend our analysis to general settings. The generalization loss for all the tasks is defined as $L(\mathcal{W})=\frac{1}{m}\sum_{i=1}^m\mathbb{E}_{(\mathbf{x},y)\sim\mathcal{D}_i}[l(f_i(\mathbf{x};\mathcal{W}),y)]$, where $\mathcal{D}_i$ denotes the underlying data distribution for the $i$th task and $\mathbb{E}[\cdot]$ denotes the expectation. The empirical loss for all the tasks is defined as $\hat{L}(\mathcal{W})=\frac{1}{m}\sum_{i=1}^m\frac{1}{n_i}\sum_{j=1}^{n_i}l(f_i(\mathbf{x}^i_j;\mathcal{W}),y^i_j)$.  We assume the loss function $l(\cdot,\cdot)$ has values in $[0,1]$ and it is Lipschitz with respect to the first input argument with a Lipschitz constant $\rho$. Each training data $\mathbf{x}^i_j$ is assumed to satisfy $\langle\mathbf{x}^i_j,\mathbf{x}^i_j\rangle\le 1$. To characterize correlations between features, we assume that $\mathbf{C}_{\mathbf{s}}=\mathbb{E}[(\mathbf{x}^i_j)_{\{\mathbf{s}\}}(\mathbf{x}^i_j)_{\{\mathbf{s}\}}^T]
\preceq\frac{\kappa}{d}\mathbf{I}$ for any $\mathbf{s}\ne\emptyset$ and $\mathbf{s}\subset[p-1]$, where $\mathbf{A}\preceq\mathbf{B}$ means that $\mathbf{B}-\mathbf{A}$ is a positive semidefinite matrix, $d=\prod_{i\in[p-1]}d_i$, and $\mathbf{I}$ denotes an identity matrix with an appropriate size.

For problem (\ref{obj_GTTN_4}), we can derive a generalization bound in the following theorem.

\begin{theorem}\label{theorem_generalization_bound}
For the solution $\hat{\mathcal{W}}$ of problem (\ref{obj_GTTN_4}) and $\delta>0$, with probability at least $1-\delta$, we have
\begin{align*}
L(\hat{\mathcal{W}})\le& \hat{L}(\hat{\mathcal{W}})+\frac{2\rho\gamma C}{mn_0}\min_{\mathbf{s}\ne\emptyset\atop\mathbf{s}\subset[p]}\left(\frac{\kappa m\sqrt{\ln d_{\mathbf{s}}}}{\alpha_{\mathbf{s}}n_0d}+\frac{\ln d_{\mathbf{s}}}{\alpha_{\mathbf{s}}n_0}\right)\\
&+\sqrt{\frac{2}{m}\ln\frac{1}{\delta}}.
\end{align*}
\end{theorem}

According to Theorem \ref{theorem_generalization_bound}, we can see that each $\alpha_{\mathbf{s}}$ can be used to weigh the second term which is related to the model complexity.

\section{Experiments}

In this section, we conduct empirical studies for the proposed GTTN.

\subsection{Experimental Settings}

\subsubsection{Datasets}

\textbf{ImageCLEF dataset}. This dataset contains 12 common categories shared by 4 tasks: Caltech-256, ImageNet ILSVRC
2012, Pascal VOC 2012, and Bing. Totally, there are about 2,400
images in all the tasks.

\textbf{Office-Caltech dataset}. This dataset consists of  4 tasks and 2,533 images in total. One task consists of data from 10 common categories shared in the Caltech-256 dataset, and the other three tasks consist of data from the Office dataset whose images are collected from
3 distinct domains/tasks, e.g., Amazon, Webcam and DSLR.

\textbf{Office-31 dataset}. This dataset contains 31 categories from Amazon, webcam, and DSLR. Totally, there are  4,110 images in all the tasks.

\textbf{Office-Home dataset}. This dataset contains images from 4 domains/tasks, which are artistic images, clip art, product
images, and real-world images. Each task contains images from 65 object categories collected in the office and home settings.
There are about 15,500 images in all the tasks.

\subsubsection{Baselines}

 We compare the GTTN method  with various competitors, including the deep multi-task learning (DMTL) method where different tasks share the first several layers as the common feature representation, the Tucker trace norm  method (denoted by Tucker), the TT trace norm method (denoted by TT), LAF trace norm method (denoted by LAF), LAF Tensor Factorisation method (denoted by LAF-TF) \cite{yh17a}.

\subsubsection{Implementation details}

\begin{figure}[!ht]
\vskip -0.1in
\begin{center}
 \includegraphics[width=\linewidth]{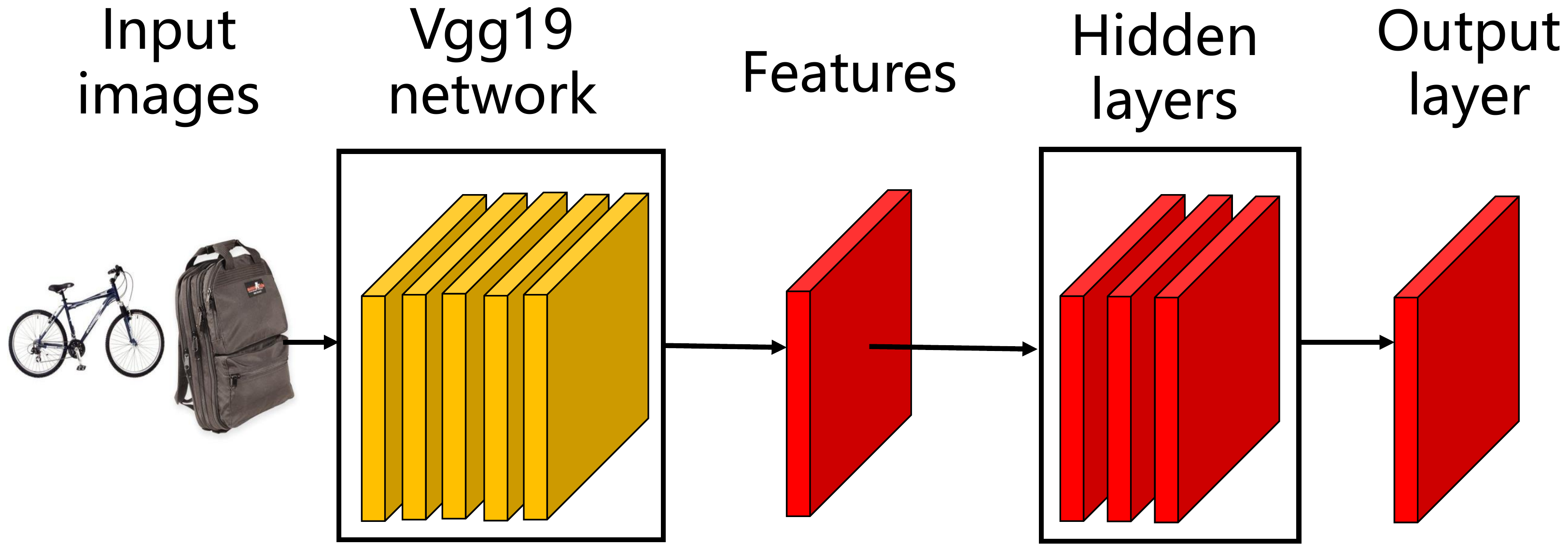}
\end{center}
\vskip -0.25in
\caption{The architecture used by all the multi-task learning models in comparison for experiments.}
\label{fig_architecture}
\vskip -0.1in
\end{figure}

We employ the Vgg19 network \cite{sz15} to extract features for image data by using the output of the pool5 layer and fc7 layer, respectively, for all the models in comparison. After that, if the pool5 layer is used, the feature representation extracted is a 3-way $7 \times 7 \times 512 $ tensor and all the multi-task learning models adopt a five-layer architecture where the three hidden layers are used to transform along each mode of the input with the ReLU activation function and they have 6, 6, 256 hidden units, respectively. Otherwise, if the fc7 layer is used, all the multi-task learning models adopt a two-layer fully-connected architecture with the ReLU activation function and 1024 hidden units, where the first layer is shared by all the tasks. The architecture used is illustrated in Figure \ref{fig_architecture}.

To see the effect of training size on the performance, we vary the training proportion from 50\% to 70\% at an interval of 10\%. The performance measure is the classification accuracy. Each experimental setting will repeat 5 times and we report the average performance as well as the standard deviation. For all the baseline methods, we follow their original model selection procedures. The regularization parameter $\lambda$ that controls the trade-off between the training cross-entropy loss and the regularization term is set by 0.25 and 0.65, respectively, for all the 6 methods to test the sensitivity of the performance with respect to to $\lambda$. In addition, we use Adam with the learning rate varying as $\eta = \frac{0.02}{1 + p}$, where $p$ is the number of the iteration and we adopt mini-batch SGD with $\text{batch\_size} = 16$.

\subsection{Experimental Results}

\begin{figure}[!ht]
\vskip -0.03in
\begin{center}
 \includegraphics[width=1\linewidth]{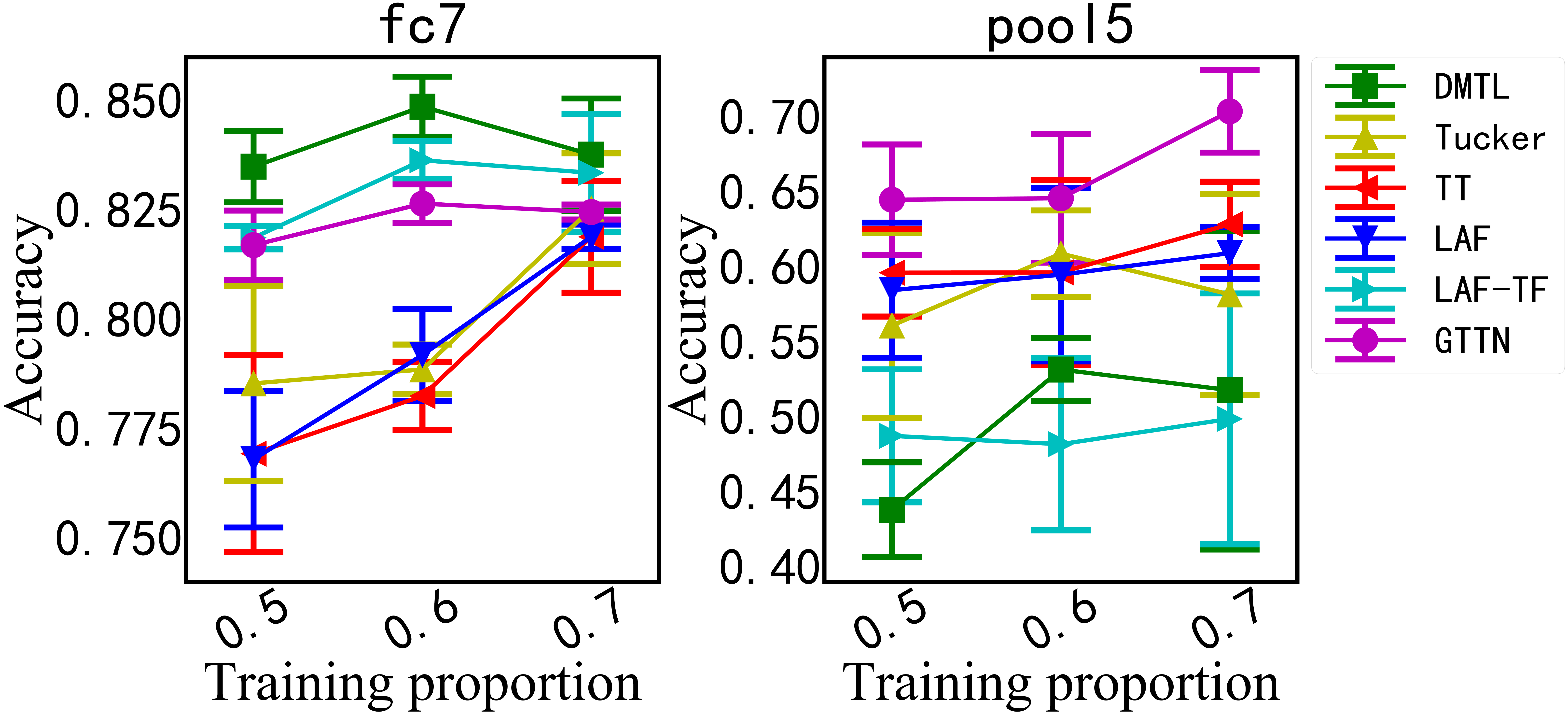}
\end{center}
\vskip -0.2in
\caption{Performance on the ImageCLEF dataset with $\lambda=0.25$.}
\label{image-CLEF-25}
\end{figure}

\begin{figure}[!ht]
\vskip -0.05in
\begin{center}
 \includegraphics[width=1\linewidth]{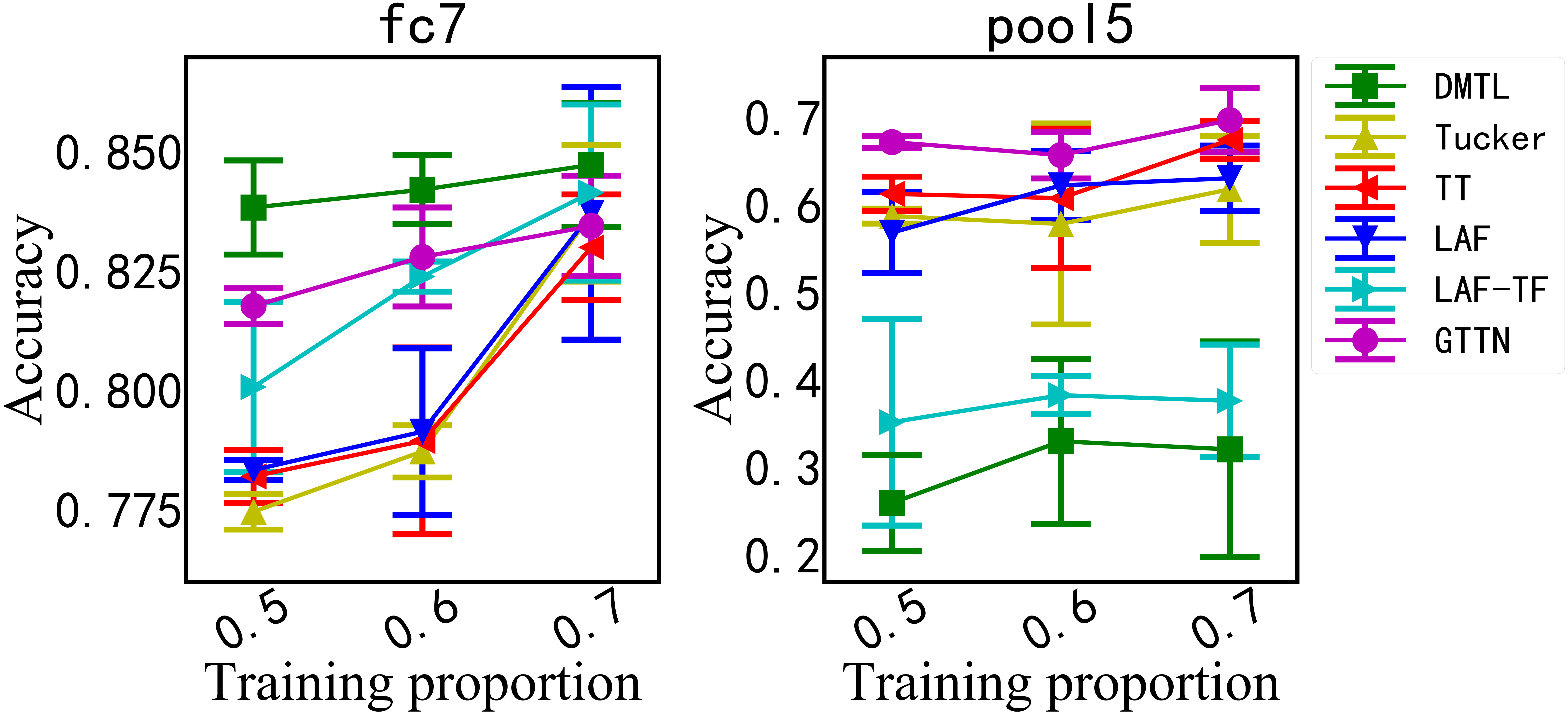}
\end{center}
\vskip -0.2in
\caption{Performance on the ImageCLEF dataset with $\lambda=0.65$.}
\label{image-CLEF-65}
\end{figure}

\begin{figure}[!ht]
\vskip -0.05in
\begin{center}
 \includegraphics[width=1\linewidth]{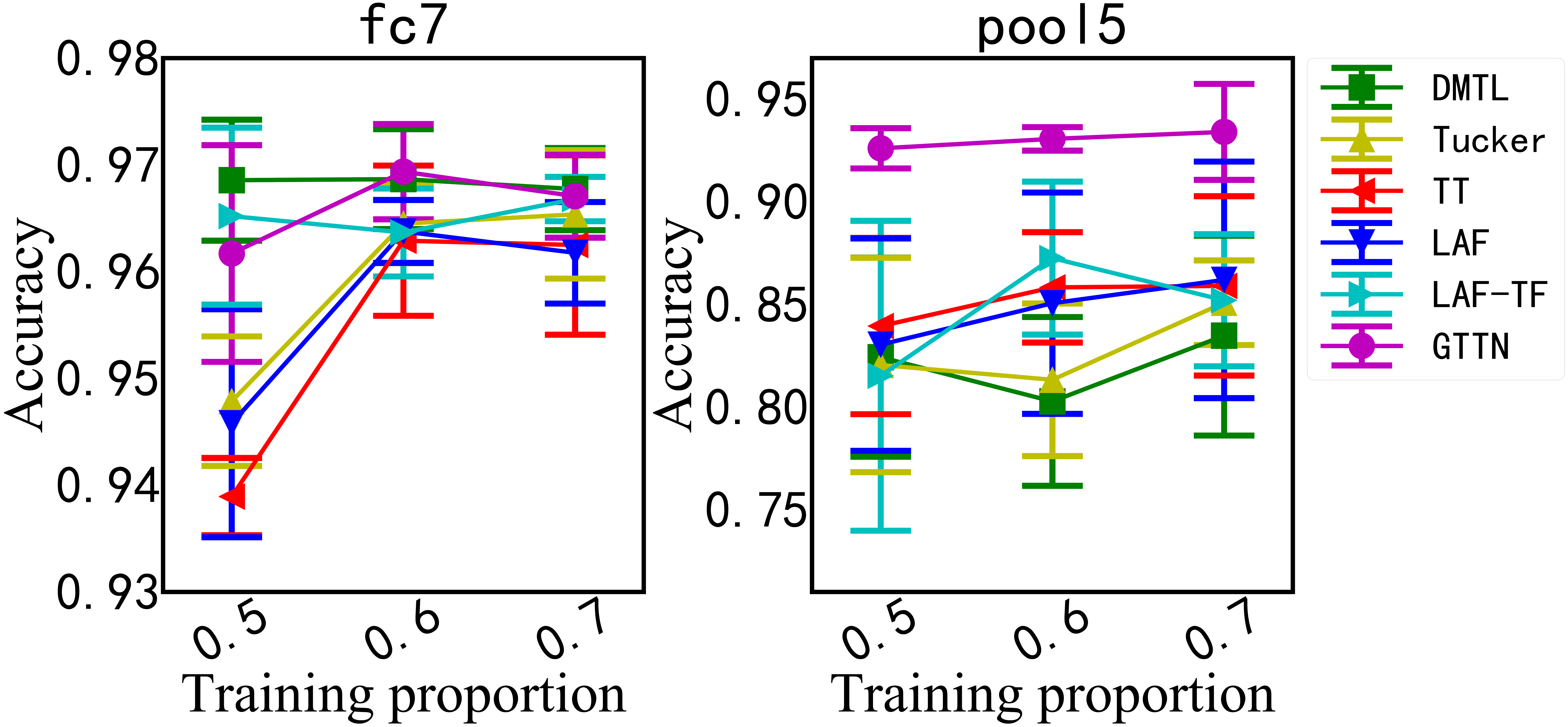}
\end{center}
\vskip -0.2in
\caption{Performance on the Office-Caltech10 dataset with $\lambda=0.25$.}
\label{Office-caltech10-25}
\end{figure}

\begin{figure}[!ht]
\vskip -0.05in
\begin{center}
 \includegraphics[width=1\linewidth]{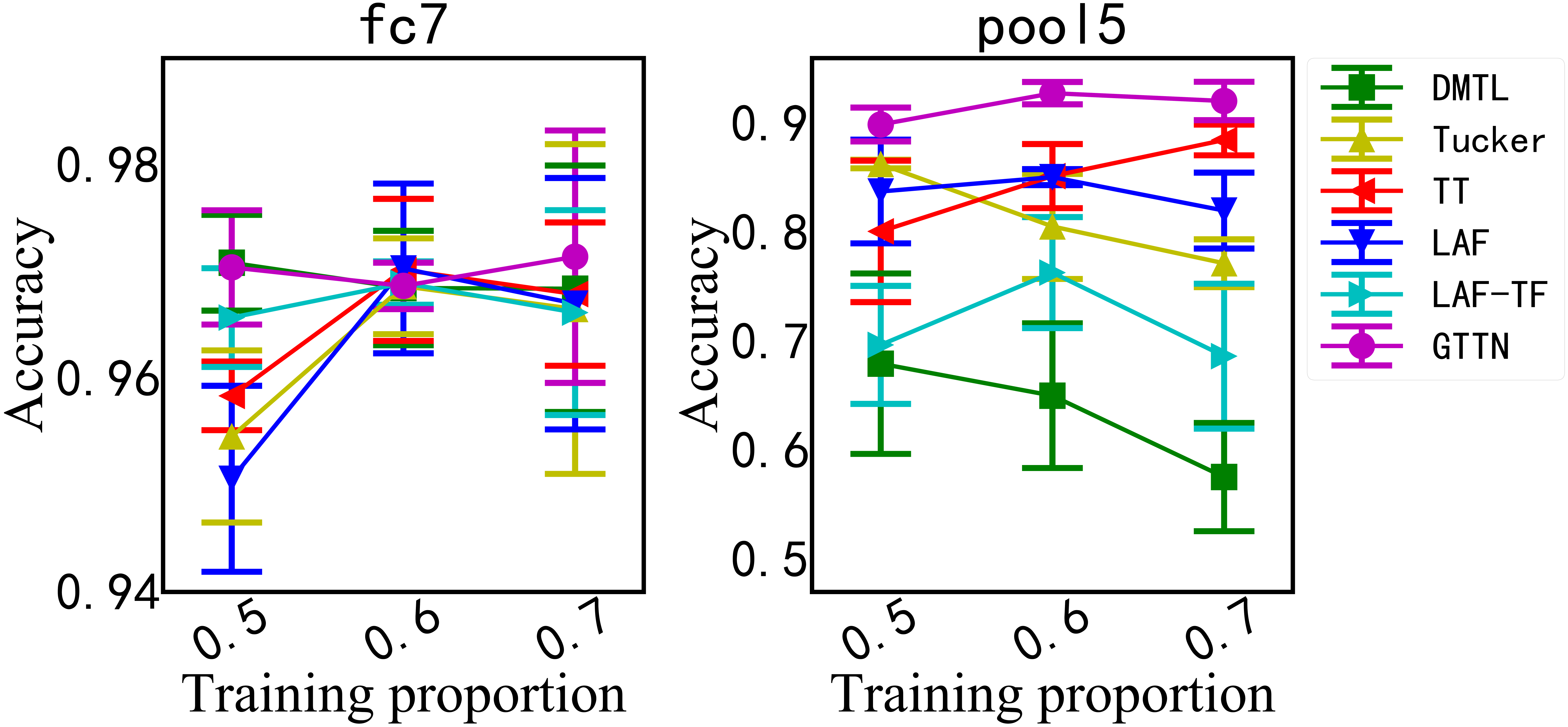}
\end{center}
\vskip -0.2in
\caption{Performance on the Office-Caltech10 dataset with $\lambda=0.65$.}
\label{Office-caltech10-65}
\end{figure}

\begin{figure}[!ht]
\vskip -0.05in
\begin{center}
 \includegraphics[width=1\linewidth]{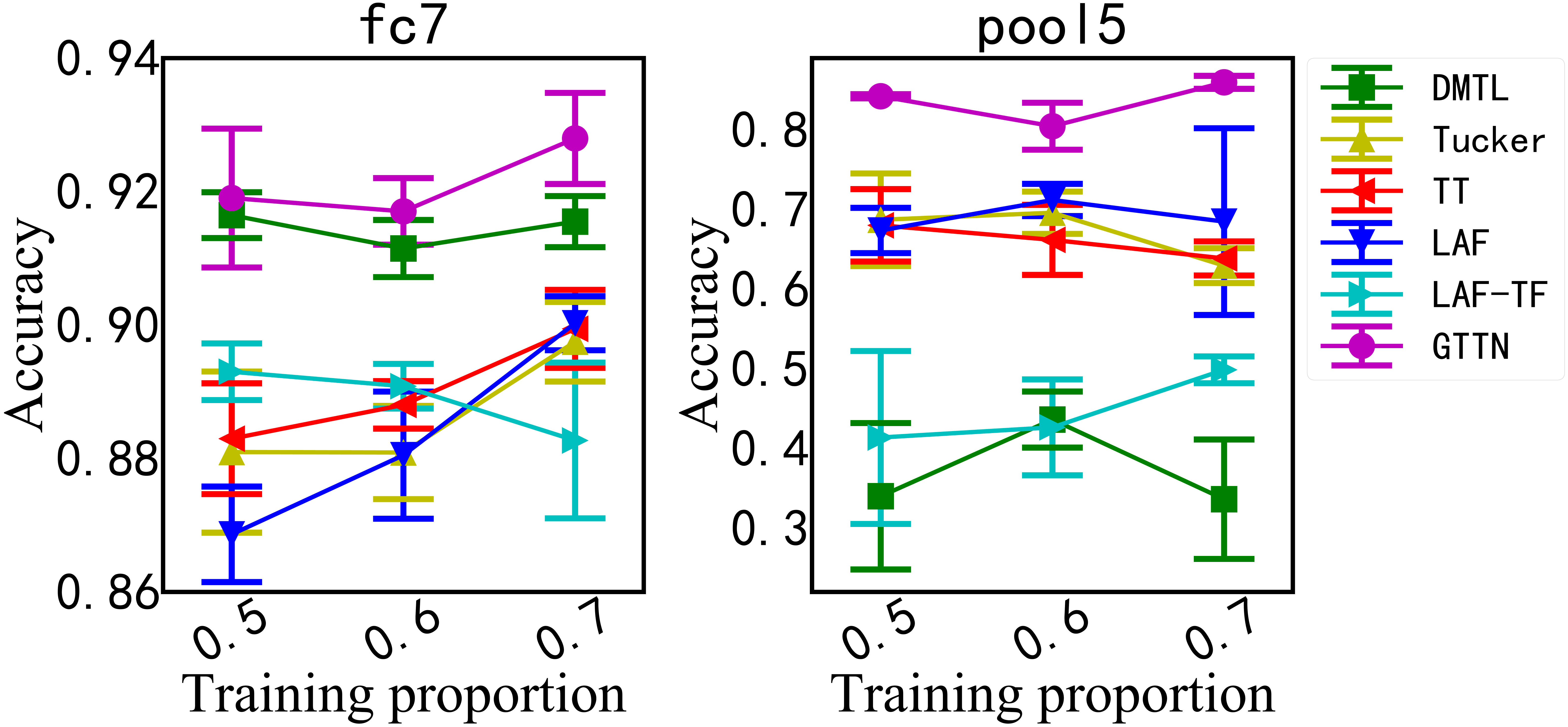}
\end{center}
\vskip -0.2in
\caption{Office-31 ($\lambda=0.25$)}
\label{Office-31-25}
\vskip -0.1in
\end{figure}

\begin{figure}[!ht]
\begin{center}
\includegraphics[width=1\linewidth]{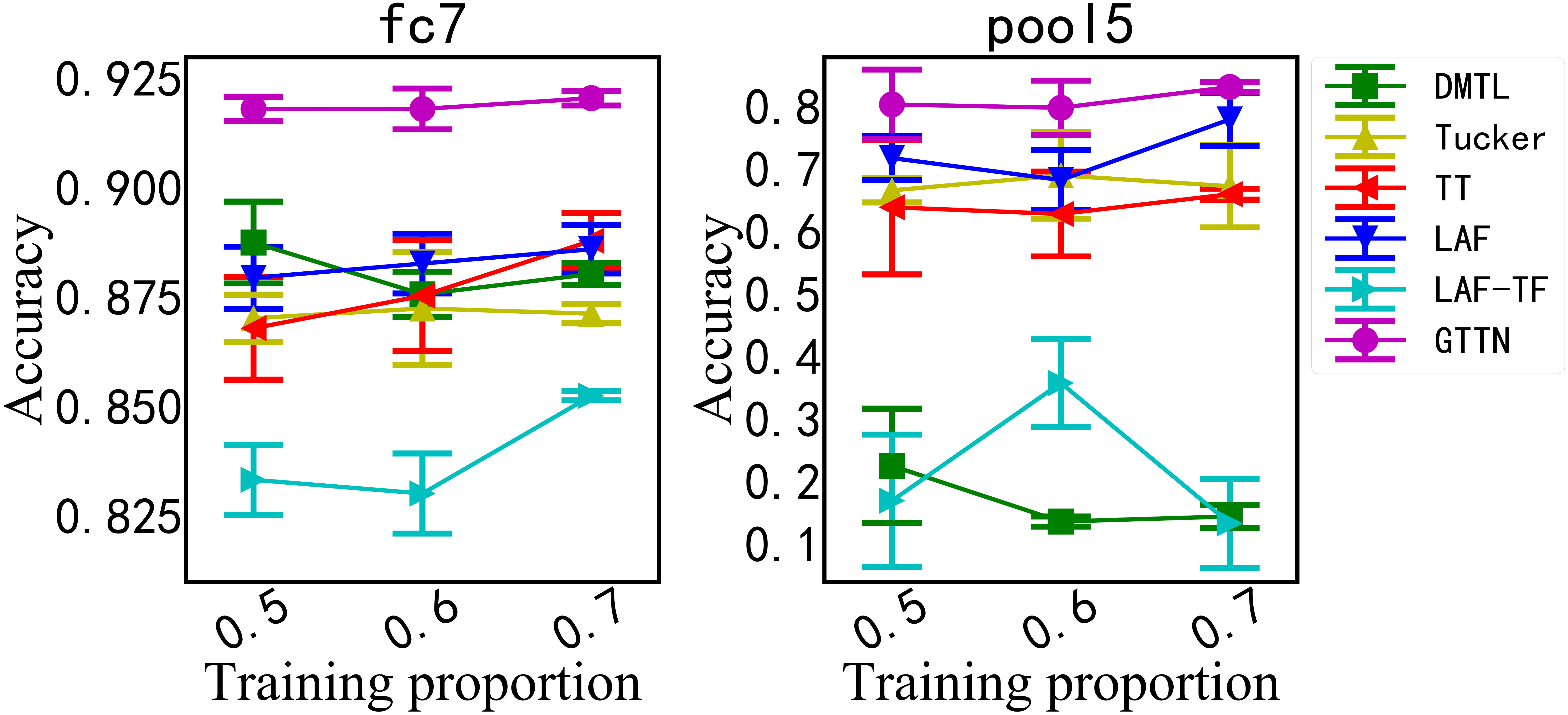}
\end{center}
\vskip -0.2in
\caption{Performance on the Office-31 dataset with $\lambda=0.65$.}
\label{Office-31-65}
\end{figure}

\begin{figure}[!ht]
\vskip -0.05in
\begin{center}
 \includegraphics[width=1\linewidth]{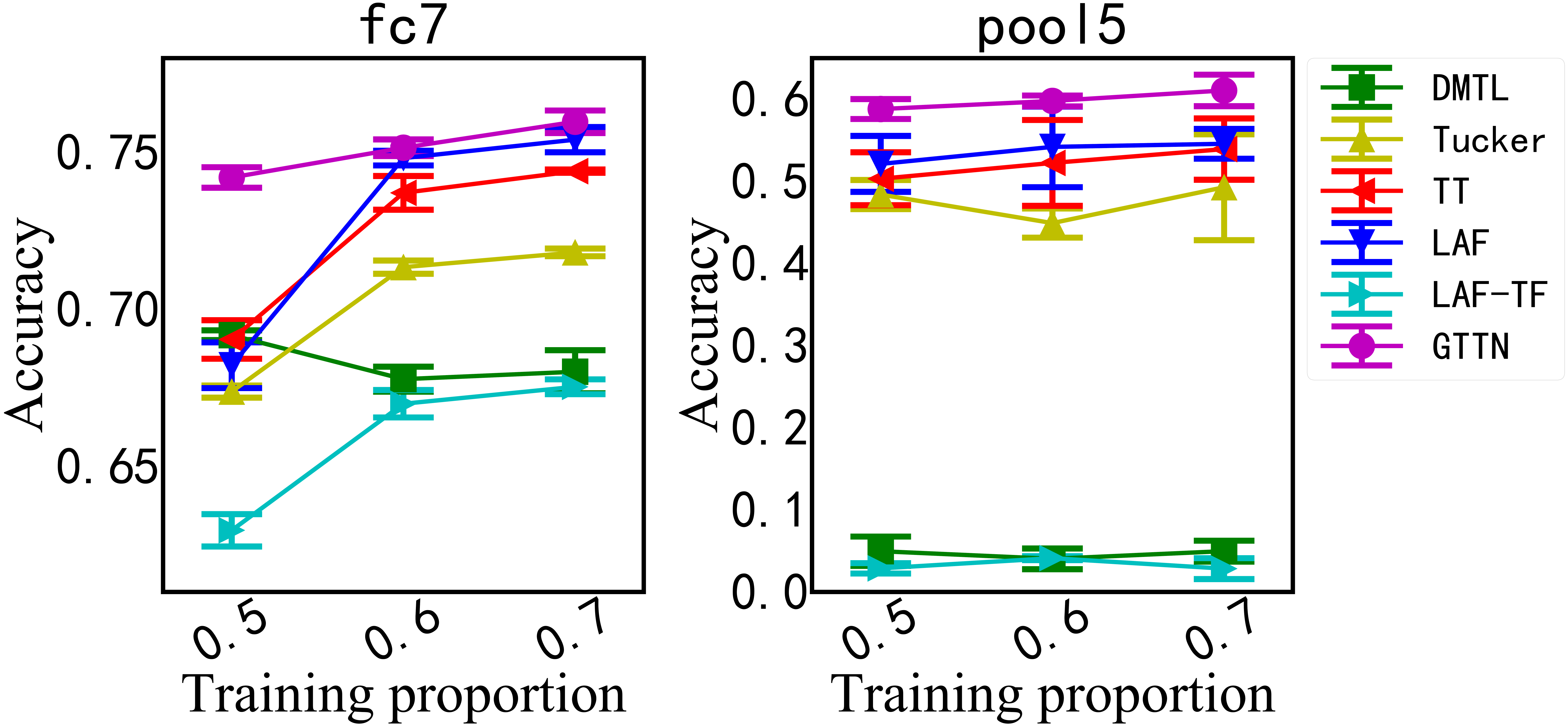}
\end{center}
\vskip -0.2in
\caption{Performance on the Office-Home dataset with $\lambda=0.25$.}
\label{Office-Home-25}
\end{figure}

\begin{figure}[!ht]
\vskip -0.05in
\begin{center}
\includegraphics[width=1\linewidth]{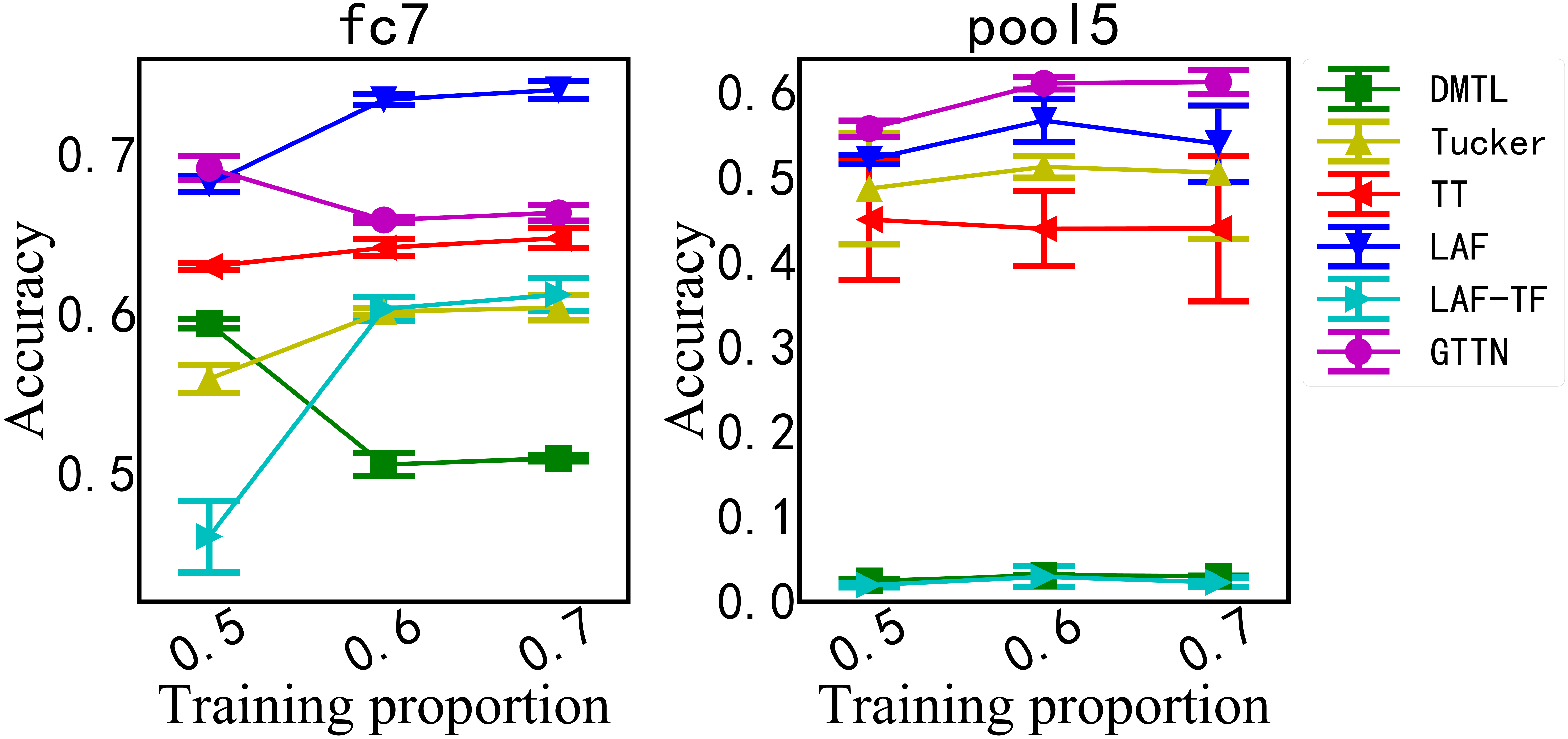}
\end{center}
\vskip -0.2in
\caption{Performance on the Office-Home dataset with $\lambda=0.65$.}
\label{Office-Home-65}
\end{figure}

The experimental results are reported in Figures \ref{image-CLEF-25}-\ref{Office-Home-65} based on different feature extractors (i.e., pool5 or fc7) and different regularization parameters (i.e., 0.25 or 0.65).

Since the output of the fc7 layer is in a vectorized representation, the model parameter $\mathcal{W}$ is a 3-way tensor. In this case, we can see that the Tucker trace norm possesses three tensor flattenings, the TT trace norm utilizes two tensor flattenings, and the GTTN also has three tensor flattenings. So in this case, both the GTTN and Tucker trace norm utilize all the possible tensor flattenings with the only difference that the GTTN learns the combination coefficients $\bm{\alpha}$ but the Tucker trace norm manually sets them to be identical. According to the results, we can see the GTTN outperforms the Tucker trace norm in most cases, which verifies that learning $\bm{\alpha}$ is better than fixing it.

When using the pool5 layer as the feature extractor, the feature representation is in a 3-way tensor, making the parameter $\mathcal{W}$ a 5-way tensor. In this case, we can see that the GTTN method performs significantly better than other baseline methods. This is mainly because the GTTN utilizes more tensor flattenings than other baseline models and hence it may discover more low-rank structures.

\subsection{Analysis on Learned $\bm{\alpha}$}

Tables \ref{pool5-25} and \ref{pool5-65} show the learned $\bm{\alpha}$ of GTTN based on the pool5 layer when $\lambda$ takes the value of 0.25 and 0.65, respectively. In this case, the parameter $\mathcal{W}$ is a 5-way tensor and hence the GTTN contains 15 different flattenings, including $\mathcal{W}_{\{1\}}$, $\mathcal{W}_{\{2\}}$, $\mathcal{W}_{\{3\}}$,  $\mathcal{W}_{\{4\}}$,
$\mathcal{W}_{\{5\}}$,
$\mathcal{W}_{\{1,2\}}$,  $\mathcal{W}_{\{1,3\}}$,  $\mathcal{W}_{\{1,4\}}$,  $\mathcal{W}_{\{2,3\}}$,  $\mathcal{W}_{\{2,4\}}$,  $\mathcal{W}_{\{3,4\}}$,  $\mathcal{W}_{\{1,2,3\}}$,  $\mathcal{W}_{\{1,2,4\}}$,  $\mathcal{W}_{\{1,3,4\}}$, and $\mathcal{W}_{\{2,3,4\}}$, which correspond to each component of $\bm{\alpha}$ in Tables  \ref{pool5-25} and \ref{pool5-65}. According to the results, we can see that different tensor flattenings have varying weights. 

Similarly, Tables  \ref{fc7-25}  and \ref{fc7-65}  show the learned $\bm{\alpha}$ of GTTN based on the fc7 layer when $\lambda$= 0.25 and $\lambda$= 0.65, respectively. In this case, the parameter $\mathcal{W}$ is a 3-way tensor, which contains 3 different flattenings by GTTN method, i.e., $\mathcal{W}_{\{1\}}$, $\mathcal{W}_{\{2\}}$ $\mathcal{W}_{\{1,2\}}$. We can notice that the weight of  ${W}_{\{1,2\}}$ is among the  maximum in most settings, which may imply that the combination of the first two axes is very important.

\begin{table*}[!htp]\small
	\centering
    \caption{\centering Learned $\bm{\alpha} $ of the GTTN with different training proportions $\theta$ (pool5, $\lambda = 0.25 $)   }
	\begin{tabular}{|c|l|l|l|}
    \hline
        \multicolumn{1}{|c|}{\bfseries Dataset  \mdseries}              & \multicolumn{1}{|c|}{\bfseries  $\bm{\alpha}$ ($\theta =0.5$ )   \mdseries}       & \multicolumn{1}{|c|}{\bfseries  $\bm{\alpha}$ ($\theta =0.6$ )   \mdseries}                &  \multicolumn{1}{|c|}{\bfseries  $\bm{\alpha}$ ($\theta =0.7$ )   \mdseries}        \\

		\hline
		\multirow{4}*{\textbf{ImageCLEF} } & 0.0736, \textbf{0.0799} , 0.0789, 0.0548,   & 0.0674, 0.0724, 0.0688, 0.0620,   &0.0757, 0.0668, 0.0699, 0.0610,   \\
		                      ~ & 0.0724, 0.0780, 0.0592, 0.0741,   & 0.0691, 0.0799, 0.0630, \textbf{0.0823},   &0.0683, \textbf{0.0819}, 0.0608, 0.0718,   \\
                             ~ & 0.0529, 0.0526, 0.0470, 0.0613,   & 0.0603, 0.0541, 0.0531, 0.0661,   & 0.0629, 0.0502, 0.0542, 0.0792,   \\
                             ~ & 0.0699, 0.0745, 0.0709  &  0.0727, 0.0657, 0.0632     &   0.0610, 0.0678, 0.0686            \\
                             \hline

		\multirow{4}*{\textbf{Office-Caltech10} } &  0.0627, 0.0739, 0.0709, 0.0604,  &  0.0722, 0.0676, 0.0783, 0.0482,   & 0.0697, 0.0762, 0.0883, 0.0497,  \\
		                      ~ &0.0707, 0.0667, 0.0564, 0.0705,   &  0.0690, 0.0725, 0.0597, 0.0705,  & \textbf{0.0901}, 0.0837, 0.0536, 0.0685,  \\
                             ~ & 0.0610, 0.0564, 0.0476, \textbf{0.0876} ,    & 0.0583, 0.0503, 0.0584, 0.0761,    &   0.0491, 0.0446, 0.0482, 0.0552,    \\
                             ~ &  0.0723, 0.0767, 0.0663   &  0.0662, \textbf{0.0842}, 0.0686    & 0.0616, 0.0768, 0.0850                   \\
                             \hline

		\multirow{4}*{\textbf{Office-31} } &  0.0796, \textbf{0.0841}, 0.0782, 0.0587,   &  0.0786, 0.0676, 0.0678, 0.0480,  &0.0778, 0.0771, \textbf{0.0805}, 0.0551,  \\
		                      ~ & 0.0771, 0.0617, 0.0577, 0.0725,  & 0.0702, \textbf{0.0843}, 0.0544, 0.0815,    & 0.0746, 0.0761, 0.0554, 0.0794,   \\
                             ~ &0.0640, 0.0557, 0.0602, 0.0571,    & 0.0578, 0.0529, 0.0651, 0.0566,   & 0.0571, 0.0510, 0.0597, 0.0489,   \\
                             ~ & 0.0505, 0.0657, 0.0771   &  0.0510, 0.0814, 0.0827     &   0.0628, 0.0705, 0.0737              \\
                             \hline

		\multirow{4}*{\textbf{Office-Home} } & \textbf{0.0867}, 0.0752, 0.0815, 0.0542,    &  0.0818, 0.0781, \textbf{0.0901}, 0.0479,   &  \textbf{0.0907}, 0.0708, 0.0784, 0.0525,\\
		                      ~ & 0.0727, 0.0831, 0.0470, 0.0798,  & 0.0872, 0.0781, 0.0522, 0.0867,  &  0.0710, 0.0795, 0.0545, 0.0848,    \\
                             ~ & 0.0550, 0.0538, 0.0810, 0.0467,   & 0.0446, 0.0451, 0.0818, 0.0439,    & 0.0517, 0.0508, 0.0744, 0.0564,   \\
                             ~ & 0.0604, 0.0480, 0.0749   &  0.0438, 0.0548, 0.0838      &  0.0617, 0.0427, 0.0802          \\
		\hline
	\end{tabular}
    \label{pool5-25}
\end{table*}

\begin{table*}[!ht]\small
\vskip -0.1in
	\centering
    \caption{\centering Learned $\bm{\alpha} $ of the GTTN with different training proportions $\theta$ (pool5, $\lambda = 0.65 $)   }
	\begin{tabular}{|c|l|l|l|}
    \hline
        \multicolumn{1}{|c|}{\bfseries Dataset  \mdseries}              & \multicolumn{1}{|c|}{\bfseries  $\bm{\alpha} $ ($\theta$ =0.5 )   \mdseries}       & \multicolumn{1}{|c|}{\bfseries  $\bm{\alpha} $ ($\theta$ =0.6 )   \mdseries}                &  \multicolumn{1}{|c|}{\bfseries  $\bm{\alpha} $ ($\theta$ =0.7 )   \mdseries}        \\

		\hline
		\multirow{4}*{\textbf{ImageCLEF} } & 0.0672, 0.0666, 0.0695, 0.0523,  & 0.0688, 0.0739, \textbf{0.0808}, 0.0602,  &\textbf{0.0821}, 0.0795, 0.0705, 0.0549,  \\
		                      ~ & 0.0712, 0.0690, 0.0670, 0.0791,  & 0.0687, 0.0680, 0.0563, 0.0726,  &0.0741, 0.0787, 0.0528, 0.0682,  \\
                             ~ & 0.0563, 0.0675, 0.0521, 0.0664,  & 0.0515, 0.0507, 0.0590, 0.0754,  & 0.0595, 0.0494, 0.0463, 0.0579,   \\
                             ~ & \textbf{0.0809}, 0.0713, 0.0637  &  0.0678, 0.0763, 0.0698     &   0.0704, 0.0743, 0.0814             \\
                             \hline

		\multirow{4}*{\textbf{Office-Caltech10} } & 0.0662, 0.0746, 0.0760, 0.0545, &  0.0681, 0.0648, \textbf{0.0863}, 0.0500,   &0.0665, 0.0730, 0.0682, 0.0613,  \\
		                      ~ & 0.0596, 0.0737, 0.0566, \textbf{0.0792},  &  0.0711, 0.0731, 0.0495, 0.0667,  & 0.0749, \textbf{0.0866}, 0.0453, 0.0857,  \\
                             ~ &0.0600, 0.0618, 0.0564, 0.0646,   & 0.0518, 0.0528, 0.0604, 0.0722,   &  0.0566, 0.0492, 0.0505, 0.0750,    \\
                             ~ &  0.0740, 0.0715, 0.0713    &  0.0721, 0.0768, 0.0841     &  0.0612, 0.0686, 0.0773               \\
                             \hline

		\multirow{4}*{ \textbf{Office-31}  } &  0.0874, 0.0772, \textbf{0.0910}, 0.0562,   &  \textbf{0.0833}, 0.0806, 0.0811, 0.0571,  &0.0680, 0.0736, 0.0788, 0.0574,   \\
		                      ~ &  0.0684, 0.0806, 0.0509, 0.0726,  & 0.0767, 0.0694, 0.0602, 0.0617,   & 0.0720, 0.0700, 0.0547, 0.0732,   \\
                             ~ &0.0518, 0.0514, 0.0621, 0.0539,   & 0.0651, 0.0575, 0.0686, 0.0553,   &  0.0535, 0.0548, 0.0622, 0.0663,   \\
                             ~ & 0.0557, 0.0642, 0.0767   &   0.0541, 0.0700, 0.0593      &   0.0588, 0.0763, \textbf{0.0804}              \\
                             \hline

		\multirow{4}*{\textbf{Office-Home} } & 0.0687, 0.0672, 0.0780, 0.0619,   &  0.0673, 0.0810, 0.0668, 0.0497,   & \textbf{0.0907}, 0.0834, 0.0835, 0.0492,  \\
		                      ~ & 0.0731, 0.0786, 0.0480, \textbf{0.0798},  & 0.0820, 0.0791, 0.0492, \textbf{0.0892} ,   & 0.0773, 0.0819, 0.0466, 0.0852,   \\
                             ~ &  0.0523, 0.0572, 0.0749, 0.0633,  & 0.0589, 0.0517, 0.0819, 0.056,   & 0.0515, 0.0432, 0.0751, 0.0522,    \\
                             ~ & 0.0591, 0.0651, 0.0730  &  0.0524, 0.0484, 0.0865     &  0.0523, 0.0523, 0.0755           \\
		\hline
	\end{tabular}
	\label{pool5-65}
\end{table*}

\begin{table*}[!ht]\small
\vskip -0.1in
\centering
\caption{\centering Learned $\bm{\alpha} $ of the GTTN with different training proportions $\theta$ (fc7, $\lambda = 0.25 $)   }
\begin{tabular}{|l|p{3cm}|p{3cm}|p{3cm}|p{3cm}|}
\hline
  \multicolumn{1}{|c|}{\bfseries Dataset  \mdseries}              & \multicolumn{1}{|c|}{\bfseries  $\bm{\alpha} $ ($\theta$ =0.5 )   \mdseries}       & \multicolumn{1}{|c|}{\bfseries  $\bm{\alpha} $ ($\theta$ =0.6 )   \mdseries}                &  \multicolumn{1}{|c|}{\bfseries  $\bm{\alpha} $ ($\theta$ =0.7 )   \mdseries}        \\

\hline
\multicolumn{1}{|c|}{\bfseries ImageCLEF    \mdseries}
              &0.3861,  0.2246, \textbf{0.3893}
              &0.3825, 0.2336,  \textbf{0.3839}
             &0.3718, 0.2154,  \textbf{0.4128}
            \\

\hline
 \multicolumn{1}{|c|}{\bfseries Office-Caltech10     \mdseries}
       &   \textbf{0.3911}, 0.2246, 0.3843
       &   \textbf{0.3953},  0.2152, 0.3895
       &   \textbf{ 0.3984}, 0.2302,0.3714           \\
\hline

 \multicolumn{1}{|c|}{\bfseries Office-31  \mdseries}
       &0.3186, 0.2507,  \textbf{0.4307}
       &0.3041, 0.2787,  \textbf{ 0.4170}
       &0.2662, 0.2864,  \textbf{ 0.4474}
       \\

\hline

\multicolumn{1}{|c|}{\bfseries Office-Home   \mdseries}
          &0.3162, 0.2750,  \textbf{0.4088}
          &0.2901, 0.2724, \textbf{0.4374}
          &0.3057, 0.2630,  \textbf{0.4313}
                 \\

\hline
\end{tabular}
\label{fc7-25}
\end{table*}

\begin{table*}[!ht]\small
\vskip -0.1in
\centering
\caption{\centering Learned $\bm{\alpha} $ of the GTTN with different training proportions $\theta$ (fc7,$\lambda = 0.65 $)}
\begin{tabular}{|l|p{3cm}|p{3cm}|p{3cm}|p{3cm}|}
\hline
  \multicolumn{1}{|c|}{\bfseries Dataset  \mdseries}              & \multicolumn{1}{|c|}{\bfseries  $\bm{\alpha} $ ($\theta$ =0.5 )   \mdseries}       & \multicolumn{1}{|c|}{\bfseries  $\bm{\alpha} $ ($\theta$ =0.6 )   \mdseries}                &  \multicolumn{1}{|c|}{\bfseries  $\bm{\alpha} $ ($\theta$ =0.7 )   \mdseries}        \\

\hline
\multicolumn{1}{|c|}{\bfseries ImageCLEF    \mdseries}
              &0.2992,  0.2834,   \textbf{0.4173}
              &0.3216, 0.2753,  \textbf{ 0.4029}
             &0.3229, 0.2908,  \textbf{0.3861}
            \\

\hline
 \multicolumn{1}{|c|}{\bfseries Office-Caltech10     \mdseries}
       &  \textbf{0.4052},  0.2244, 0.3704
       &  0.3759, 0.2462,  \textbf{0.3779}
       &  0.3871, 0.2106,  \textbf{0.4023  }           \\
\hline

 \multicolumn{1}{|c|}{\bfseries Office-31  \mdseries}
       &0.3609, 0.2184, \textbf{ 0.4207}
       & \textbf{ 0.3926}, 0.2415, 0.3658
       &0.3279, 0.2399,  \textbf{0.4322}
       \\

\hline

\multicolumn{1}{|c|}{\bfseries Office-Home   \mdseries}
          &0.2789,  0.2944,  \textbf{0.4267}
          &0.3113,  0.2618,  \textbf{0.4269}
          &0.2746,  0.2672,   \textbf{0.4582}
                 \\

\hline
\end{tabular}
\label{fc7-65}
\end{table*}

\section{Conclusion}

In this paper, we devise a generalized tensor trace norm to capture all the low-rank structures in a parameter tensor used in deep multi-task learning and identify the importance of each structure. We analyze properties of the proposed GTTN, including its dual norm and generalization bound. Empirical studies show that it outperforms state-of-the-art counterparts and the learned combination coefficients can give us more understanding of the problem studied. As a future work, we are interested in extending the idea of GTTN to study tensor Schatten norms.



\bibliographystyle{icml2020}
\bibliography{GTTN}

\section*{Appendix}

\subsection*{Proof for Theorem \ref{theorem_num_summands}}

{\bf Proof}. For a valid $\|\mathcal{W}_{\{\mathbf{s}\}}\|_*$, it is required that $\mathbf{s}$ and $\neg\mathbf{s}$ should not be empty, implying that $\mathbf{s}\ne \emptyset$ and $\mathbf{s}\ne [p]$. So the total number of valid summands in the right-hand side of Eq. (\ref{GTTN}) is $2^p-2$. Based on the definition of $\mathcal{W}_{\{\mathbf{s}\}}$, we can see that $\mathcal{W}_{\{\mathbf{s}\}}$ is equal to the transpose $\mathcal{W}_{\{\neg\mathbf{s}\}}$, making $\|\mathcal{W}_{\{\mathbf{s}\}}\|_*=\|\mathcal{W}_{\{\neg\mathbf{s}\}}\|_*$. So for $\|\mathcal{W}_{\{\mathbf{s}\}}\|_*$, there will always be an equivalent $\|\mathcal{W}_{\{\neg\mathbf{s}\}}\|_*$, leading to $2^{p-1}-1$ distinct summands in the right-hand side of Eq. (\ref{GTTN}). \hfill$\Box$

\subsection*{Proof for Theorem \ref{theorem_obj_GTTN_reformulation}}

{\bf Proof}. Based on Eq. (\ref{GTTN}), we rewrite problem (\ref{obj_GTTN}) as
\begin{equation*}
\min_{\bm{\Theta},\bm{\alpha}\in\mathcal{C}_{\bm{\alpha}}}\sum_{i=1}^m\frac{1}{n_i}\sum_{j=1}^{n_i}l(f_i(\mathbf{x}^i_j;\bm{\Theta}),y^i_j)
+\lambda\sum_{\mathbf{s}\subset[p]\atop\mathbf{s}\ne\emptyset}\alpha_{\mathbf{s}}\|\mathcal{W}_{\{\mathbf{s}\}}\|_*,
\end{equation*}
which is equivalent to
\begin{equation*}
\min_{\bm{\Theta}}\sum_{i=1}^m\frac{1}{n_i}\sum_{j=1}^{n_i}l(f_i(\mathbf{x}^i_j;\bm{\Theta}),y^i_j)
+\lambda\min_{\bm{\alpha}\in\mathcal{C}_{\bm{\alpha}}}
\sum_{\mathbf{s}\subset[p]\atop\mathbf{s}\ne\emptyset}\alpha_{\mathbf{s}}\|\mathcal{W}_{\{\mathbf{s}\}}\|_*.
\end{equation*}
So we just need to prove that $$\min_{\bm{\alpha}\in\mathcal{C}_{\bm{\alpha}}}
\sum_{\mathbf{s}\subset[p]\atop\mathbf{s}\ne\emptyset}\alpha_{\mathbf{s}}\|\mathcal{W}_{\{\mathbf{s}\}}\|_*=
\min_{{\mathbf{s}\subset[p]\atop\mathbf{s}\ne\emptyset}}\|\mathcal{W}_{\{\mathbf{s}\}}\|_*.$$
The optimization problem in the left-hand side of the above equation is a linear programming problem with \mbox{respect} to $\bm{\alpha}$. It is easy to show that $\sum_{\mathbf{s}\subset[p]}\alpha_{\mathbf{s}}\|\mathcal{W}_{\{\mathbf{s}\}}\|_*\ge \min_{\mathbf{s}\subset[p]\atop\mathbf{s}\ne\emptyset}\|\mathcal{W}_{\{\mathbf{s}\}}\|_*$ for $\bm{\alpha}\in\mathcal{C}_{\bm{\alpha}}$, where the equality holds when the corresponding coefficient for $\min_{\mathbf{s}\subset[p]\atop\mathbf{s}\ne\emptyset}\|\mathcal{W}_{\{\mathbf{s}\}}\|_*$ equals 1 and other coefficients equals 0. Then we reach the conclusion.\hfill$\Box$

\subsection*{Proof for Theorem \ref{theorem_GTTN_dual_norm}}

{\bf Proof}. 
We define a linear operator $\Phi(\mathcal{W})=[\mathrm{vec}(\alpha_{\{[1]\}}\mathcal{W}_{\{[1]\}});\ldots;\alpha_{\{[2:p]\}}\mathrm{vec}(\mathcal{W}_{\{[2:p]\}})]$, where $\mathrm{vec}(\cdot)$ denotes the columnwise concatenation of a matrix and $[i:j]$ denotes a set of successively integers for $i$ to $j$. We define the $q$ norm as
\begin{equation*}
\|\mathbf{y}\|_q=\sum_i\|\mathcal{Y}^{(\pi(i))}_{\{\pi(i)\}}\|_*,
\end{equation*}
where $\mathcal{Y}^{(\pi(i))}_{\{\pi(i)\}}$ denotes the inverse vectorization of a subvector $\mathbf{z}_{(i-1)*N+1:kN}$ of $\mathbf{z}$ into a $\prod_{j\in\pi(i)}p_j\times\prod_{j\in\neg\pi(i)}p_j$ matrix where $N=\prod_{j=1}^pd_j$ and $\pi(i)$ transforms an index $i$ into a subset of $[p]$. Based on the definition of the dual norm, we have
\begin{equation*}
\vertiii{\mathcal{W}}_{*^{\star}}=\sup\langle\mathcal{W},\mathcal{X}\rangle\ \mathrm{s.t.}\ \vertiii{\mathcal{X}}_{*}\le 1,
\end{equation*}
where $\langle\cdot,\cdot\rangle$ denotes the inner product between two tensors with equal size. Since this maximization problem satisfies the Slater condition, the strong duality holds. Thus, due to Fenchel duality theorem, we have
{\small
\begin{equation*}
\sup_{\mathcal{X}}(\langle\mathcal{W},\mathcal{X}\rangle-\delta(\vertiii{\mathcal{X}}_{*}\le 1))=\inf_{\mathbf{y}}(\delta(-\Phi^T(\mathbf{y})+\mathcal{X})+\|\mathbf{y}\|_{q^\star}),
\end{equation*}
}\noindent
where $\delta(C)$ is an indicator function of condition $C$ and it outputs 0 when $C$ is true and otherwise $\infty$. Since the dual norm of the trace norm is the spectral norm, we reach the conclusion.\hfill$\Box$

\subsection*{Proof for Theorem \ref{theorem_generalization_bound}}

Before presenting the proof for Theorem \ref{theorem_generalization_bound}, we first prove the following theorem.

\begin{theorem}\label{theorem_M_upperbound}
$\sigma^i_j$, a Rademacher variable, is an uniform $\{\pm 1\}$-valued random variable, and $\mathcal{M}$ is a $d_1\times\ldots\times d_{p-1}\times d_p$ tensor with $\mathcal{M}_i=\sum_{j=1}^{n_0}\frac{1}{n_0}\sigma^i_j\mathbf{x}^i_j$, where $d_p$ equals $m$. Then we have
\begin{equation*}
\mathbb{E}[\vertiii{\mathcal{M}}_{*^\star}]\le \min_{\mathbf{s}\ne\emptyset\atop\mathbf{s}\subset[p]}\frac{C}{\alpha_{\mathbf{s}}}\left(\frac{\kappa m}{n_0d}\sqrt{\ln d_{\mathbf{s}}}+\frac{\ln d_{\mathbf{s}}}{n_0}\right).
\end{equation*}
where $d_{\mathbf{s}}=\prod_{i\in\mathbf{s}}d_i+\prod_{j\in\neg\mathbf{s}}d_j$, $C$ is an absolute constant,
\end{theorem}
{\bf Proof}. We define $d_p=m$. According to Theorem \ref{theorem_GTTN_dual_norm}, we have
\begin{equation*}
\vertiii{\mathcal{M}}_{*^\star}=\min_{\sum_{\mathbf{s}\ne\emptyset\atop\mathbf{s}\subset[p]}\alpha_{\mathbf{s}}\mathcal{Y}^{(\mathbf{s})}=\mathcal{M}}
\max_{\mathbf{s}\ne\emptyset\atop\mathbf{s}\subset[p]}\|\mathcal{Y}^{(\mathbf{s})}_{\{\mathbf{s}\}}\|_{\infty}
\end{equation*}
Since for each $\mathbf{s}$ we can make $\alpha_{\mathbf{s}}\mathcal{Y}^{\mathbf{s}}$ equal to $\mathcal{M}$, we have
\begin{equation*}
\vertiii{\mathcal{M}}_{*^\star}\le \frac{1}{\alpha_{\mathbf{s}}}\|\mathcal{M}_{\{\mathbf{s}\}}\|_{\infty}\ \forall \mathbf{s}\ne\emptyset,\ \mathbf{s}\subset[p],
\end{equation*}
which implies that
\begin{equation*}
\vertiii{\mathcal{M}}_{*^\star}\le \min_{\mathbf{s}}\frac{1}{\alpha_{\mathbf{s}}}\|\mathcal{M}_{\{\mathbf{s}\}}\|_{\infty}.
\end{equation*}
So we can get
\begin{align*}
\mathbb{E}[\vertiii{\mathcal{M}}_{*^\star}]
\le&\mathbb{E}\left[\min_{\mathbf{s}}\frac{1}{\alpha_{\mathbf{s}}}\|\mathcal{M}_{\{\mathbf{s}\}}\|_{\infty}\right]\\
\le&\min_{\mathbf{s}}\mathbb{E}\left[\frac{1}{\alpha_{\mathbf{s}}}\|\mathcal{M}_{\{\mathbf{s}\}}\|_{\infty}\right].
\end{align*}

Based on Theorem 6.1 in \cite{tropp12}, we can upper-bound each expectation as
\begin{align*}
\mathbb{E}\left[\|\mathcal{M}_{\{\mathbf{s}\}}\|_{\infty}\right]\le C(\sigma_{\mathbf{s}}\sqrt{\ln d_{\mathbf{s}}}+\psi_{\mathbf{s}}\ln d_{\mathbf{s}}),
\end{align*}
where $\mathcal{Z}^{i,j}$ is a $d_1\times\ldots\times d_{p-1}\times d_p$ zero tensor with only the $i$th slice along the last axis equal to $\frac{1}{n_0}\sigma^i_j\mathbf{x}^i_j$, $\psi_{\mathbf{s}}$ needs to satisfy $\psi_{\mathbf{s}}\ge\|\mathcal{Z}^{i,j}_{\{\mathbf{s}\}}\|_{\infty}$, and
{\scriptsize
\begin{align*}
&\sigma_{\mathbf{s}}^2\\
=&\max\Big(
\big\|\sum_{i=1}^{m}\sum_{j=1}^{n_0}\mathbb{E}\big[\mathcal{Z}^{i,j}_{\{\mathbf{s}\}}(\mathcal{Z}^{i,j}_{\{\mathbf{s}\}})^T\big]\big\|_{\infty},
\big\|\sum_{i=1}^{m}\sum_{j=1}^{n_0}\mathbb{E}\big[(\mathcal{Z}^{i,j}_{\{\mathbf{s}\}})^T\mathcal{Z}^{i,j}_{\{\mathbf{s}\}}\big]\big\|_{\infty}
\Big).
\end{align*}
}\noindent
As the Frobenius norm of a matrix is larger than its spectral norm, $\|\mathcal{Z}^{i,j}_{\{\mathbf{s}\}}\|_{\infty}\le \frac{1}{n_0}$ and we simply set $\psi_{\mathbf{s}}=\frac{1}{n_0}$. For $\sigma_{\mathbf{s}}$, we have
\begin{equation*}
\mathbb{E}\Big[\sum_{j=1}^{n_0}\mathcal{Z}^{i,j}_{\{\mathbf{s}\}}(\mathcal{Z}^{i,j}_{\{\mathbf{s}\}})^T\Big]=\frac{1}{n_0}
\mathbf{C}_{\mathbf{s}-\{p\}}\preceq\frac{\kappa}{n_0d}\mathbf{I},
\end{equation*}
implying that
\begin{equation*}
\left\|\sum_{i=1}^{m}\sum_{j=1}^{n_0}\mathbb{E}\big[\mathcal{Z}^{i,j}_{\{\mathbf{s}\}}(\mathcal{Z}^{i,j}_{\{\mathbf{s}\}})^T\big]\right\|_{\infty}
\le\frac{\kappa m}{n_0d}.
\end{equation*}
Similarly, we have
\begin{equation*}
\mathbb{E}\Big[\sum_{j=1}^{n_0}(\mathcal{Z}^{i,j}_{\{\mathbf{s}\}})^T\mathcal{Z}^{i,j}_{\{\mathbf{s}\}}\Big]
=\mathrm{diag}\left(\frac{\mathrm{tr}(\mathbf{C}_{\mathbf{s}-\{p\}})}{n_0}\right)\preceq\frac{\kappa}{n_0d}\mathbf{I},
\end{equation*}
where $\mathrm{tr}(\cdot)$ denotes the trace of a matrix and $\mathrm{diag}(\cdot)$ converts a vector or scalar to a diagonal matrix. This inequality implies
\begin{equation*}
\left\|\sum_{i=1}^{m}\sum_{j=1}^{n_0}\mathbb{E}\big[\mathcal{Z}^{i,j}_{\{\mathbf{s}\}}(\mathcal{Z}^{i,j}_{\{\mathbf{s}\}})^T\big]\right\|_{\infty}
\le\frac{\kappa m}{n_0d}.
\end{equation*}
By combining the above inequalities, we reach the conclusion.\hfill$\Box$

Then we can prove Theorem \ref{theorem_generalization_bound} as follows.

{\bf Proof}. By following \cite{bm02}, we have
\begin{eqnarray*}
L(\hat{\mathcal{W}})&\le& \hat{L}(\hat{\mathcal{W}})+\sup_{\vertiii{\mathcal{W}}_*\le\gamma}\left\{L(\mathcal{W})-\hat{L}(\mathcal{W})\right\}\\
&=&\hat{L}(\hat{\mathcal{W}})+\sup_{\vertiii{\mathcal{W}}_*\le\gamma}\left\{\mathbb{E}[\hat{L}(\mathcal{W})]-\hat{L}(\mathcal{W})\right\}.
\end{eqnarray*}
When each pair of the training data $(\mathbf{x}^i_j,y^i_j)$ changes, the random variable $\sup_{\vertiii{\mathcal{W}}_*\le\gamma}\left\{\mathbb{E}[\hat{L}(\mathcal{W})]-\hat{L}(\mathcal{W})\right\}$ can change by no more than $\frac{2}{mn_0}$ due to the boundedness of the loss function $l(\cdot,\cdot)$. Then by McDiarmid's inequality, we can get
{\scriptsize
\begin{align*}
&P\left(\sup_{\mathcal{W}\in\mathcal{C}}\left\{\mathbb{E}[\hat{L}(\mathcal{W})]-\hat{L}(\mathcal{W})\right\}
-\mathbb{E}\left[\sup_{\mathcal{W}\in\mathcal{C}}\left\{\mathbb{E}[\hat{L}(\mathcal{W})]-\hat{L}(\mathcal{W})\right\}\right]\ge t\right)\\
&\le \exp\left\{-\frac{t^2mn_0}{2}\right\},
\end{align*}
}\noindent
where $P(\cdot)$ denotes the probability and $\mathcal{C}=\{\mathcal{W}|\vertiii{\mathcal{W}}_*\le\gamma\}$. This inequality implies that with probability at least $1-\delta$,
{\small
\begin{align*}
\sup_{\mathcal{W}\in\mathcal{C}}\left\{\mathbb{E}[\hat{L}(\mathcal{W})]-\hat{L}(\mathcal{W})\right\}
\le&\mathbb{E}\left[\sup_{\mathcal{W}\in\mathcal{C}}\left\{\mathbb{E}[\hat{L}(\mathcal{W})]-\hat{L}(\mathcal{W})\right\}\right]\\
&+\sqrt{\frac{2}{mn_0}\ln\frac{1}{\delta}}.
\end{align*}
}\noindent
Based on the the property of the Rademacher complexity, we have
\begin{align*}
&\mathbb{E}\left[\sup_{\mathcal{W}\in\mathcal{C}}\left\{\mathbb{E}[\hat{L}(\mathcal{W})]-\hat{L}(\mathcal{W})\right\}\right]\\\le&
2\rho\mathbb{E}\left[\sup_{\mathcal{W}\in\mathcal{C}}\left\{\frac{1}{mn_0}\sum_{i=1}^m\sum_{j=1}^{n_0}\sigma^i_jf_i(\mathbf{x}^i_j)\right\}\right].
\end{align*}
Then based on the definition of $\mathcal{M}$ and the H\"{o}lder's inequality, we have
\begin{equation*}
\sup_{\mathcal{W}\in\mathcal{C}}\left\{\frac{1}{mn_0}\sum_{i=1}^m\sum_{j=1}^{n_0}\sigma^i_jf_i(\mathbf{x}^i_j)\right\}
\le\frac{\gamma}{m}\vertiii{\mathcal{M}}_{*^\star}.
\end{equation*}
By combining the above inequalities, with probability at least $1-\delta$, we have
\begin{align*}
L(\hat{\mathcal{W}})&\le& \hat{L}(\hat{\mathcal{W}})+\frac{2\rho\gamma}{m}\mathbb{E}[\vertiii{\mathcal{M}}_{*^\star}]+\sqrt{\frac{2}{mn_0}\ln\frac{1}{\delta}}.
\end{align*}
Then by incorporating Theorem \ref{theorem_M_upperbound} into this inequality, we reach the conclusion.\hfill$\Box$

\end{document}